\documentclass{article}
\usepackage{iclr2027_conference,times}
\usepackage{hyperref}
\usepackage{url}
\usepackage{graphicx}
\usepackage{subcaption}
\usepackage{wrapfig}
\usepackage{xspace}
\usepackage{algorithm}
\usepackage{algpseudocode}
\usepackage{booktabs}
\usepackage{multirow}
\usepackage{adjustbox}
\usepackage{makecell}
\usepackage{arydshln}
\usepackage{placeins}
\usepackage{amsmath}
\usepackage{amssymb}
\newcommand{\methodname}{\textsc{Forge}\xspace}
\title{Beyond Generator-Specific Shortcuts: Open-World AI-Generated Video Detection via Adversarial Training}
\author{Qi Zhou\textsuperscript{$\dagger$}, Yuanfan Li\textsuperscript{$\dagger$}, Huiqi Deng\textsuperscript{$\ast$}, Hongbin Pei, Yazhe Tang \\
  Faculty of Electronic and Information Engineering, Xi'an Jiaotong University \\
  \textsuperscript{$\dagger$} Equal contribution; \textsuperscript{$\ast$} Correspondence to: \texttt{denghq7@xjtu.edu.cn}}
\iclrfinalcopy

\begin{document}
\maketitle
\begin{abstract}
AI-generated video (AIGV) detection is increasingly important for mitigating misinformation, impersonation, and other malicious uses of synthetic media, yet existing detectors often generalize poorly to unseen generators in open-world settings. To better understand this challenge, we analyze the failure modes of AIGV detectors under cross-generator out-of-distribution (OOD) shifts and observe that detector representations strongly encode generator-specific cues, which are associated with substantial ID--OOD performance drops. Further analysis shows that naively perturbing generator-sensitive tokens can even reduce OOD performance, indicating that generator-related cues and detection-relevant evidence are only partially separable. Motivated by these findings, we propose Forensic Obfuscation of geneRator-specific cues via Gradient-guided Embedding perturbation (\methodname), a token-aware adversarial training framework for open-world AIGV detection. \methodname\ identifies tokens that are strongly associated with generator-specific shortcuts yet less critical to real/fake discrimination, and selectively perturbs them during training to suppress generator-dependent patterns while preserving useful forensic evidence. Through synchronized adversarial training between an attacker and a detector, \methodname\ encourages the detector to learn more transferable generator-invariant forensic representations. Experiments on 29 held-out benchmark entries show that \methodname\ achieves an average F1 of 92.62, outperforming eight strong baselines by 5.31 points, while also remaining robust under four common post-processing attacks. We will release our code and datasets upon acceptance.

\end{abstract}

\section{Introduction}

Models such as Sora2~\cite{brooks2024video} and Veo~\cite{collins2024new} now produce high-quality, realistic AI-generated videos (AIGVs) at scale. These advances benefit content creation, education, and entertainment, but also enable large-scale production of misleading or harmful videos. Such videos can be exploited for fake news, impersonation, fraud, and large-scale disinformation, often without reliable attribution. Researchers have therefore developed automated AIGV detectors to help platforms and users distinguish real from AI-generated (fake) videos and identify their possible origin.

\begin{figure}[htbp]
\centering
    \captionsetup[subfigure]{justification=centering}
    \begin{subfigure}[t]{0.32\linewidth}
        \centering
        \includegraphics[width=\linewidth]{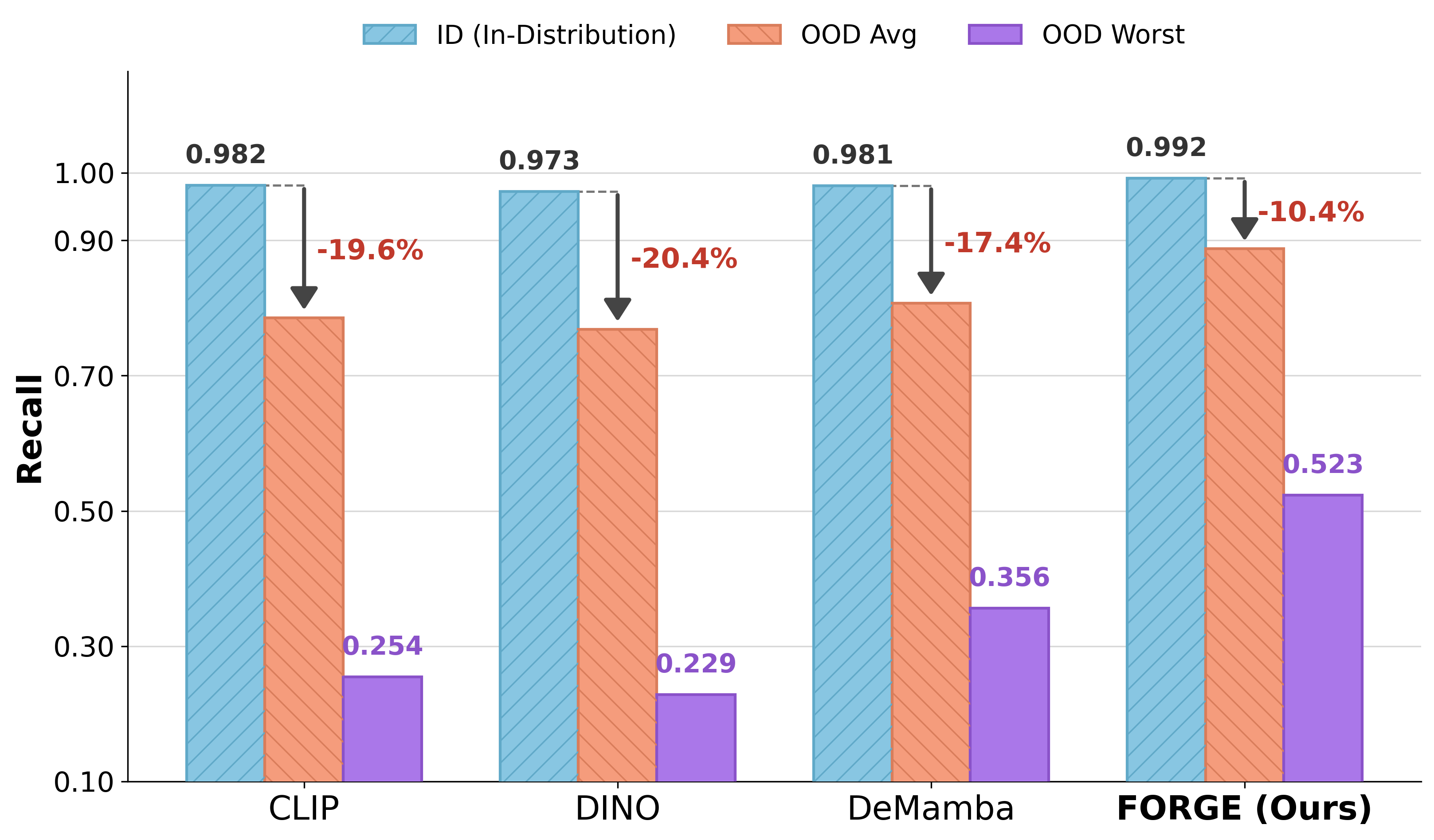}
        \caption{Recall Drop on Held-Out Entries}
        \label{fig:recall_drop}
    \end{subfigure}%
    \hfill
    \begin{subfigure}[t]{0.35\linewidth}
        \centering
        \includegraphics[width=\linewidth]{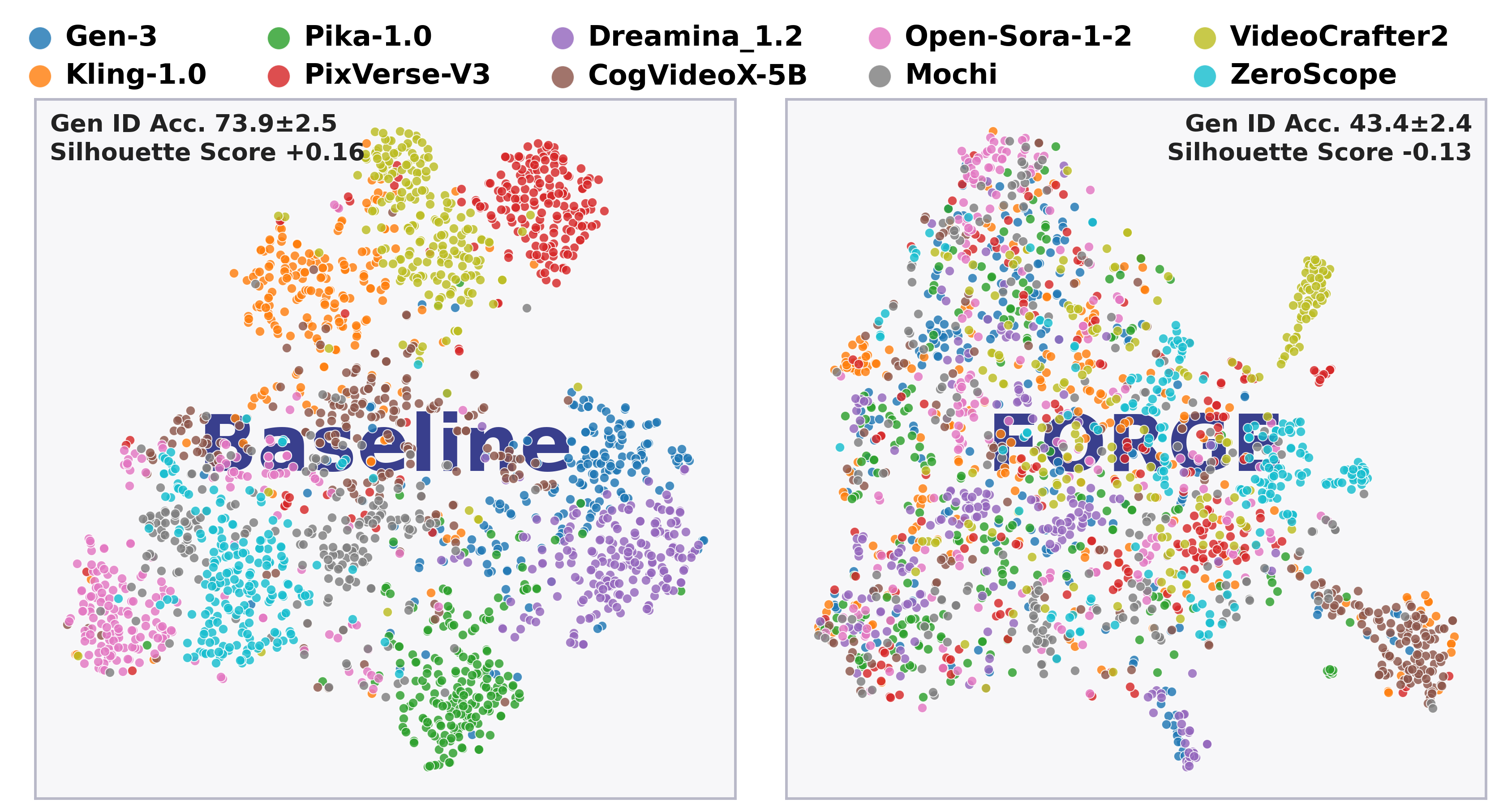}
        \caption{t-SNE Results of Detectors}
        \label{fig:tsne}
    \end{subfigure}%
    \hfill
    \begin{subfigure}[t]{0.29\linewidth}
        \centering
        \includegraphics[width=\linewidth]{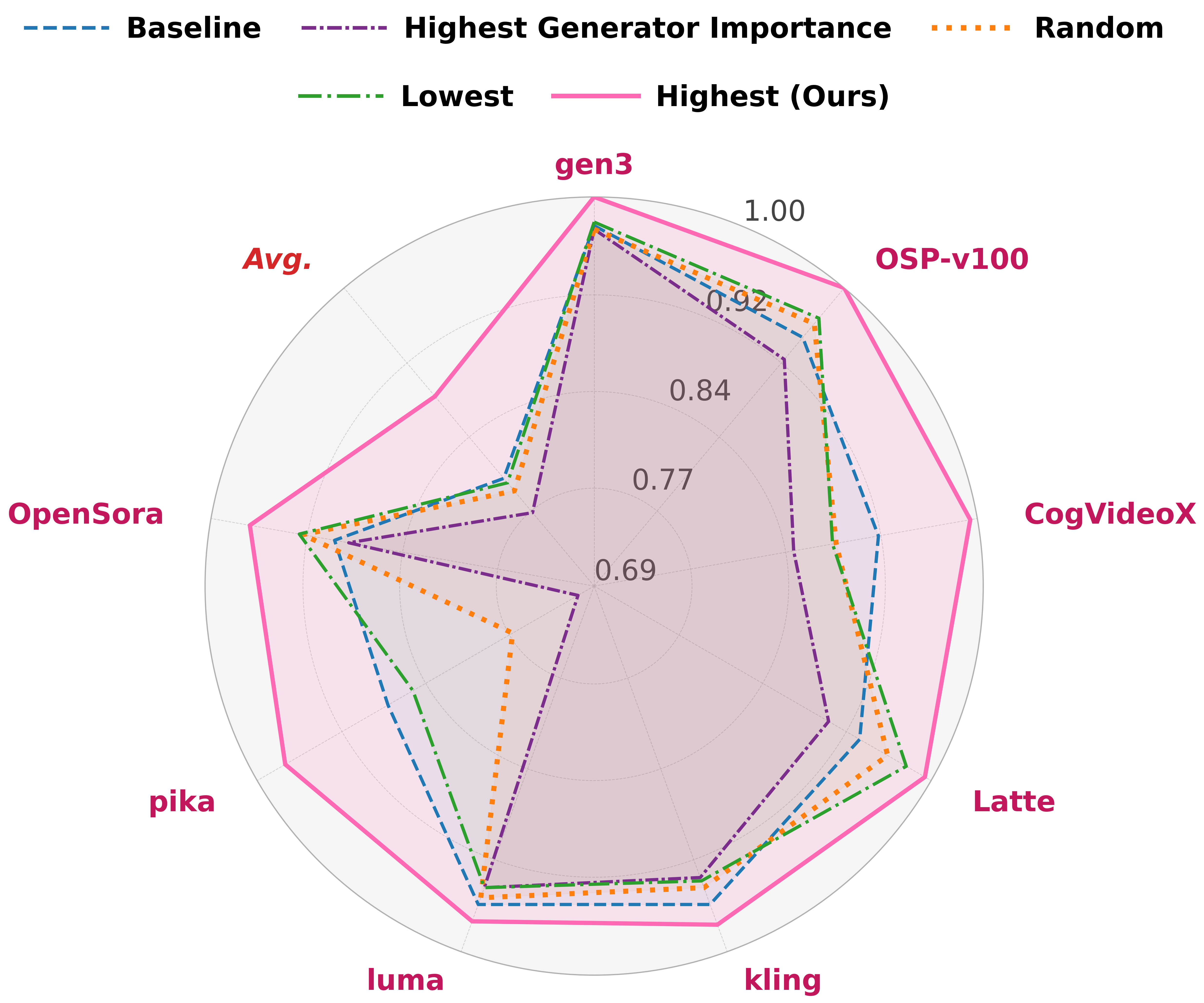}
        \caption{Token Perturbation Study}
        \label{fig:modify}
    \end{subfigure}
    \caption{\textbf{Analysis of generalization degradation and token selection.}
    \textbf{(a)} Detector recall drops substantially on held-out test entries.
    \textbf{(b)} t-SNE shows that baseline embeddings are highly separable by generator, indicating reliance on generator-specific shortcut cues, whereas our \methodname mitigates this effect.
    \textbf{(c)} We compare four token perturbation strategies: \textbf{Random}, \textbf{Lowest}, \textbf{Highest Generator Importance}, and \textbf{Ours}. Perturbing highly generator-sensitive tokens harms recall, while random or low-importance perturbation does not improve OOD generalization. Our method achieves the best generalization performance.}
    \label{fig:intro}
\end{figure}

Existing AIGV detectors~\cite{zheng2025d3,chen2024demamba,decof2025,unite2025} still generalize poorly to unseen generators in the rapidly evolving AIGV landscape. Figure~\ref{fig:intro}(a) shows a clear recall drop for all detectors on held-out test entries, limiting their practical reliability.

This performance drop under held-out evaluation is also consistent with recent findings in related AI-generated content detection tasks~\cite{zheng2024breaking,corvi2025seeing,han2025towards}, where OOD degradation is often attributed to detectors relying on semantic artifacts, dataset shortcuts, or other model-specific biases instead of transferable forensic evidence. The substantial gap between training and held-out entries suggests that current detectors may not primarily rely on generator-invariant forensic traces, but instead exploit cues tied to the generators observed during training. Motivated by this observation, we analyze whether the learned detector representations retain information predictive of generator identity. As shown in Figure~\ref{fig:intro}(b), we freeze the trained detector and train a lightweight linear probe for generator classification using only \textbf{5} samples per generator across the 10 seen generators. Even in this extremely low-shot setting, the baseline detector still reaches a high probe accuracy of $73.9 \pm 2.5$, and its embeddings exhibit clear generator-wise clustering with a silhouette coefficient of $0.16$. Such strong few-shot separability suggests that generator identity is strongly encoded in the learned representation. In contrast, a detector trained with our design reduces the probe accuracy to $43.4 \pm 2.4$ and the silhouette coefficient to $-0.13$, while improving OOD recall by around 10 points. This evidence suggests that strong generator-dependent encoding is associated with weaker cross-generator generalization.

To test whether generator-dependent cues can be suppressed directly, we perform the token perturbation analysis in Figure~\ref{fig:intro}(c), where each token is characterized by two complementary notions: \emph{generator relevance}, measuring its contribution to generator identification, and \emph{detection relevance}, measuring its contribution to real/fake discrimination. However, directly perturbing the top-ranked tokens selected solely by generator relevance reduces OOD recall, indicating that generator-sensitive tokens cannot be naively treated as removable shortcuts. Motivated by this observation, we propose \methodname, a token-aware adversarial training framework for open-world AIGV detection. Instead of perturbing tokens based only on generator importance, \methodname\ selectively perturbs those that are highly informative for generator identity yet less critical to the detector's decision, thereby suppressing generator-specific shortcuts while preserving detection-relevant evidence. Combined with generator-invariant representation learning, this design encourages the detector to rely less on generator-dependent cues and more on transferable forensic traces, leading to stronger OOD generalization than random perturbation or generator-only selection.

\textbf{Our Work.}
In this paper, we propose \textbf{F}orensic \textbf{O}bfuscation of gene\textbf{R}ator-specific cues via \textbf{G}radient-guided \textbf{E}mbedding perturbation (\methodname), an adversarial training framework for open-world AIGV detection. \methodname\ comprises two interacting modules: an attacker and a detector, optimized jointly throughout training. The attacker aims to identify and obfuscate generator-specific shortcuts by selectively perturbing tokens that are strongly associated with generator identity but less critical to the detector's real/fake decision. This targeted design suppresses generator-dependent cues while preserving detection-relevant evidence to the greatest extent possible. The detector is then trained on both clean and perturbed videos, driving it away from shortcut reliance and toward more transferable, generator-invariant forensic traces. Experiments on 29 held-out benchmark entries demonstrate that \methodname\ achieves an F1 score of 92.62, surpassing 8 state-of-the-art detectors by 5.31 points, while maintaining robustness against four attack methods targeting AIGV detectors. Our contributions are as follows:

\begin{itemize}
    \setlength{\itemsep}{0pt}
    \setlength{\parsep}{0pt}
    \setlength{\topsep}{0pt}
    \item \textbf{Token-aware Adversarial Training Framework.}
    We propose \methodname, a token-aware adversarial training framework to improve the open-world generalization of AIGV detectors. In this framework, an attacker identifies and perturbs generator-sensitive tokens to expose generator-specific shortcuts, while the detector is trained on both clean and perturbed videos to preserve real/fake discriminative ability and learn more transferable forensic traces.

    \item \textbf{Generator-sensitive Token Identification.}
    We propose a selective token identification strategy that jointly considers generator relevance and detection relevance. By combining signals from an auxiliary generator classifier and the detector itself, our method identifies tokens that are strongly associated with generator-specific cues yet less critical to the detection decision, enabling more precise suppression of generator-dependent shortcuts without destroying useful forensic evidence.

    \item \textbf{Outstanding Performance.}
    Experimental results on 29 held-out benchmark entries show that \methodname achieves an F1 score of 92.62, outperforming 8 SOTA detectors by 5.31 points. Moreover, \methodname remains robust against four common attack methods targeting AIGV detectors, demonstrating stronger open-world generalization and attack robustness.
\end{itemize}

\section{Related Work}
\textbf{Open-world AI-Generated Video (AIGV) Detection}
With the rapid development of AI-generated video models, a growing number of AIGV detection methods have been proposed~\cite{chen2024demamba,zheng2025d3,zhang2025physics}. These methods exploit diverse visual and temporal cues to distinguish real videos from AI-generated ones, and have achieved strong performance in standard in-distribution settings~\cite{interno2025ai,bai2024ai}.
However, despite their success, many existing detectors suffer substantial performance drops in open-world OOD scenarios, especially on unseen generators. This limits their practical applicability, since real-world AIGV systems evolve rapidly and new generators emerge continuously.
Different from these methods, our approach explicitly improves cross-generator generalization. Through adversarial training, we identify and perturb generator-specific features that are not essential for detection, encouraging the detector to learn more generator-invariant forensic cues and thereby improving OOD robustness in open-world AIGV detection.

\noindent\textbf{Adversarial Training.}
Adversarial training is a widely used strategy for improving model robustness by exposing the model to hard perturbations or adversarial examples during training. Recent studies~\cite{wang2022deepfake,khan2024adversarially,li2025iron,li2026learning} have shown its effectiveness across multiple modalities, including AI-generated text detection and deepfake detection, where adversarially constructed samples are used to enhance detector robustness under attacks. However, most existing works mainly use adversarial training to improve robustness against input-space perturbations or attack-time evasion. In contrast, our method uses adversarial training for a different purpose: improving open-world OOD generalization in AIGV detection. Instead of simply defending against external attacks, our attacker explicitly identifies and perturbs generator-specific yet detection-safe tokens, forcing the detector to reduce its reliance on generator-dependent shortcuts and learn more transferable forensic cues.
\section{Methodology}
\begin{figure}[htbp]
    \centering
    \resizebox{1\textwidth}{!}{%
        \includegraphics{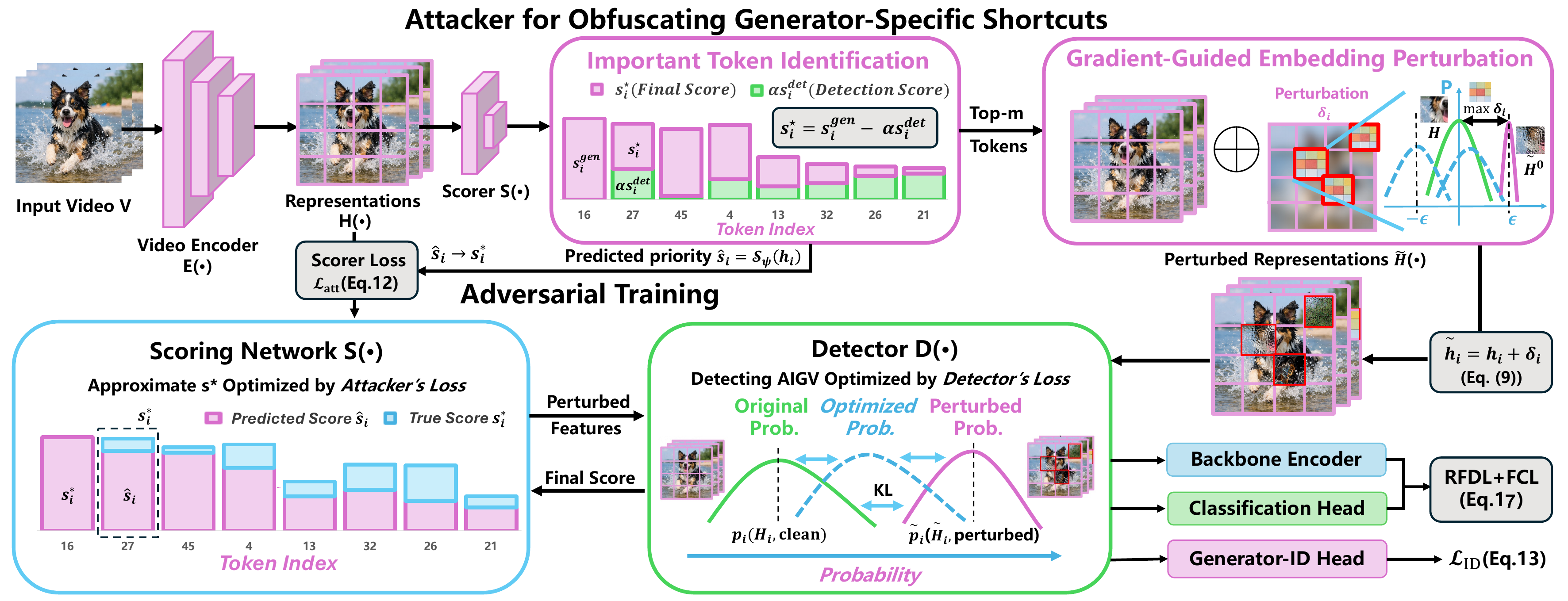} 
    }

    \caption{\textbf{Pipeline of \methodname.}
    The attacker identifies important AIGV tokens, applies targeted gradient-guided embedding perturbations, and feeds the resulting perturbed representations to the detector for adversarial training. The detector objective combines RFDL and FCL as defined in Eq.~(\ref{eq:l_det_total}).}
    \label{fig:framework}
\end{figure}

\subsection{Attacker for Obfuscating Generator-Specific Shortcuts}
\label{sec:attacker}
The attacker uses a learnable scoring network \(\mathcal{S}_{\psi}(\cdot)\) to obfuscate generator-specific shortcuts during training through important-token identification and gradient-guided embedding perturbation.

\noindent \textbf{Important Token Identification.}
As shown in Figure~\ref{fig:intro}(b), detector representations strongly encode generator identity, and such generator-dependent encoding is associated with weaker OOD generalization. Therefore, the first role of the attacker is to identify the tokens that carry generator-specific cues and make them the primary targets for perturbation.

Given an input video \(V\), we first randomly sample \(k\) frames from it and feed them into a ViT-based video encoder \(\mathcal{E}(\cdot)\) to obtain \(s\) spatiotemporal token representations:
\begin{equation}
\mathbf{H} = \mathcal{E}(V) = [h_1, h_2, \dots, h_s], \qquad h_i \in \mathbb{R}^d,
\label{eq:att_hidden}
\end{equation}
where \(h_i\) is the representation of the \(i\)-th token.

To identify the tokens that are most suitable for perturbation, we follow prior work~\cite{simonyan2013deep,selvaraju2017grad} in using gradient magnitude as a measure of token importance, where larger gradients indicate that the target objective is more sensitive to the corresponding token representation.
We first quantify each token's relevance to generator identity in order to locate generator-specific cues. To this end, we introduce an auxiliary generator classifier and define the generator classification loss as
\begin{equation}
\mathcal{L}_{\mathrm{gen}} = - \log p_{\mathrm{gen}}(y_{\mathrm{gen}} \mid V),
\label{eq:l_gen}
\end{equation}
where \(y_{\mathrm{gen}}\) is the generator label. The generator relevance score of the \(i\)-th token is then computed as
\begin{equation}
s_i^{\mathrm{gen}} = \left\| \frac{\partial \mathcal{L}_{\mathrm{gen}}}{\partial h_i} \right\|_2 .
\label{eq:s_gen}
\end{equation}
A larger \(s_i^{\mathrm{gen}}\) indicates that the token contributes more strongly to generator identification and is therefore more likely to encode generator-specific cues.

However, Figure~\ref{fig:intro}(c) shows that generator importance alone is insufficient for token selection, since directly perturbing the most generator-sensitive tokens already reduces OOD recall. This suggests that some generator-sensitive tokens also contain useful evidence for real/fake discrimination. Therefore, besides locating generator-specific cues, we also need to estimate how critical each token is to the detector itself. Let
\begin{equation}
\mathcal{L}_{\mathrm{det}} = - \log p_{\mathrm{det}}(y \mid V),
\label{eq:l_det}
\end{equation}
where \(y\) is the real/fake label. We define the detection relevance score of the \(i\)-th token as
\begin{equation}
s_i^{\mathrm{det}} = \left\| \frac{\partial \mathcal{L}_{\mathrm{det}}}{\partial h_i} \right\|_2 .
\label{eq:s_det}
\end{equation}
A larger \(s_i^{\mathrm{det}}\) means that the token is more important to the detector's decision and should therefore be perturbed more cautiously.

To jointly account for these two factors, we define the final perturbation priority score as
\begin{equation}
s_i^{\star} = s_i^{\mathrm{gen}} - \alpha s_i^{\mathrm{det}},
\label{eq:s_final}
\end{equation}
where \(\alpha\) is a balancing coefficient. Under this formulation, tokens with high generator relevance but low detection relevance receive higher priority for perturbation, since they are more likely to encode generator-specific shortcuts while being less essential for real/fake discrimination.

We use the gradient-based scores in Eqs.~(\ref{eq:s_gen})--(\ref{eq:s_final}) to supervise a learnable scoring network \(\mathcal{S}_{\psi}(\cdot)\). During adversarial training, these targets are computed for generated videos in the current batch, while the network predicts a priority from each token representation:
\begin{equation}
\hat{s}_i = \mathcal{S}_{\psi}(h_i).
\label{eq:s_hat} 
\end{equation}
The scoring network is trained through the \emph{Attacker's Loss} to track the refined score \(s_i^{\star}\) as the detector evolves. Its predicted scores \(\hat{s}_i\) determine which tokens are perturbed: we rank the tokens and select the top-\(m\) positions:
\begin{equation}
\mathcal{I} = \operatorname{TopK}\left(\{(\hat{s}_i, h_i)\}_{i=1}^{s},\, m\right).
\label{eq:topk}
\end{equation}

\noindent\textbf{Gradient-Guided Embedding Perturbation.}
After identifying the important token set \(\mathcal{I}\), we apply targeted perturbations to the selected token embeddings in order to obfuscate generator-specific cues while preserving the overall detection semantics. Given the token representations \(\mathbf{H} = [h_1, h_2, \dots, h_s]\), we construct the perturbed representations \(\tilde{\mathbf{H}} = [\tilde{h}_1, \tilde{h}_2, \dots, \tilde{h}_s]\) as
\begin{equation}
\tilde{h}_i = h_i + \mathbf{1}[i \in \mathcal{I}] \, \delta_i,
\label{eq:perturb_embed}
\end{equation}
where \(\delta_i\) is the perturbation applied to the \(i\)-th token, and \(\mathbf{1}[i \in \mathcal{I}]\) is an indicator function that equals \(1\) if \(i \in \mathcal{I}\), and \(0\) otherwise.

To compute \(\delta_i\), we first initialize the perturbation from a uniform distribution and normalize it:
\begin{equation}
\delta_i^{0} \sim U(a,b), \qquad 
\hat{\delta}_i^{0} = \xi \frac{\delta_i^{0}}{\|\delta_i^{0}\|_2},
\label{eq:delta_init}
\end{equation}
where \(\xi\) is a scaling factor, and \(a\) and \(b\) are the lower and upper bounds of the uniform distribution. Based on the initialized perturbation, we form an initial perturbed representation \(\tilde{\mathbf{H}}^{0}\).

We then adopt a single-step gradient-guided update to move the perturbation towards the direction that most strongly alters the generator classifier's output distribution. Specifically, the perturbation is computed as
\begin{equation}
\delta_i
=
\epsilon
\frac{
\nabla_{\hat{\delta}_i^{0}}
\mathrm{KL}
\big(
p_{\mathrm{gen}}(\cdot \mid \mathbf{H})
\;\|\;
p_{\mathrm{gen}}(\cdot \mid \tilde{\mathbf{H}}^{0})
\big)
}{
\left\|
\nabla_{\hat{\delta}_i^{0}}
\mathrm{KL}
\big(
p_{\mathrm{gen}}(\cdot \mid \mathbf{H})
\;\|\;
p_{\mathrm{gen}}(\cdot \mid \tilde{\mathbf{H}}^{0})
\big)
\right\|_2
},
\label{eq:delta_update}
\end{equation}
where \(\epsilon\) controls the perturbation magnitude, and \(p_{\mathrm{gen}}(\cdot \mid \mathbf{H})\) and \(p_{\mathrm{gen}}(\cdot \mid \tilde{\mathbf{H}}^{0})\) denote the generator classifier output distributions before and after perturbation, respectively.

In this way, the perturbation is explicitly guided toward directions that maximally disrupt generator-discriminative information in the selected tokens. The resulting perturbed token representations \(\tilde{\mathbf{H}}\) are then fed into the detector together with the clean representations for subsequent adversarial training.

\noindent \textbf{Attacker's Loss.}
For each input video \(V_j\) in a batch of size \(N\), we obtain its token representations
\(\mathbf{H}_j = [h_{j,1}, h_{j,2}, \dots, h_{j,s}]\) through the video encoder \(\mathcal{E}(\cdot)\).
Based on Eqs.~(\ref{eq:s_gen})--(\ref{eq:s_final}), we compute the refined perturbation priority
\(s_{j,i}^{\star}\) for each token \(h_{j,i}\), and use it as the supervision target for the scoring network.
Let \(\hat{s}_{j,i} = \mathcal{S}_{\psi}(h_{j,i})\) denote the score predicted by the attacker. The attacker loss is then defined as
\begin{equation}
\mathcal{L}_{\mathrm{att}}
=
\frac{1}{Ns}
\sum_{j=1}^{N}
\sum_{i=1}^{s}
\left(
\hat{s}_{j,i} - s_{j,i}^{\star}
\right)^2 .
\label{eq:l_att}
\end{equation}

Minimizing Eq.~(\ref{eq:l_att}) enables the scoring network to fit the refined token priority that jointly reflects generator relevance and detection relevance, providing more reliable token selection for subsequent perturbation. In each training step, we update the scoring network using \(\mathcal{L}_{\mathrm{att}}\) while keeping the detector fixed. The alternating optimization is described later in this section.

\subsection{Detector for Learning Generator-Invariant Forensic Traces}

Our detector \(\mathcal{D}(\cdot)\) consists of a shared backbone encoder \(\mathcal{E}(\cdot)\), a real/fake classification head \(C_{\mathrm{det}}(\cdot)\), and a generator-ID classification head \(C_{\mathrm{gen}}(\cdot)\). During training, it takes as input the clean representations \(\mathbf{H}_i=\mathcal{E}(V_i)\) and the perturbed representations \(\tilde{\mathbf{H}}_i\) generated by the attacker. The goal of the detector is to preserve accurate real/fake predictions under perturbation while reducing reliance on generator-dependent shortcuts and learning more transferable generator-invariant forensic traces. To this end, we optimize the detector with a task-preserving classification loss and a forensic consistency loss.

\noindent \textbf{Task-Preserving Classification Loss.}
The Task-Preserving Classification Loss (TCL) maintains both reliable generator supervision for the attacker and accurate real/fake discrimination for the detector. The generator-ID head provides stable generator-aware supervision for token-level generator relevance estimation in Sec.~\ref{sec:attacker}. For an input video \(V_i\) with generator label \(g_i\), we optimize the generator-ID head on the clean representations \(\mathbf{H}_i\) using
\begin{equation}
\mathcal{L}_{\mathrm{ID}}
=
\frac{1}{N}
\sum_{i=1}^{N}
-\log p_{\mathrm{gen}}(g_i \mid \mathbf{H}_i),
\label{eq:l_id}
\end{equation}
where \(N\) is the batch size and \(p_{\mathrm{gen}}(g_i \mid \mathbf{H}_i)\) is the probability assigned by \(C_{\mathrm{gen}}(\cdot)\) to the correct generator label.

For real/fake detection, let \(y_i \in \{0,1\}\) be the label of \(V_i\), and write \(q_i=p_{\mathrm{det}}(\mathbf{H}_i)\) and \(\tilde q_i=p_{\mathrm{det}}(\tilde{\mathbf{H}}_i)\). The shared backbone and the real/fake head are optimized on both clean and perturbed representations via the \textbf{Real/Fake Detection Loss (RFDL)}:
\begin{equation}
\begin{aligned}
\mathcal{L}_{\mathrm{RFDL}}
=
-\frac{1}{N}\sum_{i=1}^{N}
\Big[
& y_i \log q_i + (1-y_i)\log(1-q_i) \\
& + y_i \log \tilde q_i + (1-y_i)\log(1-\tilde q_i)
\Big],
\end{aligned}
\label{eq:l_rfdl}
\end{equation}
where \(q_i\) and \(\tilde q_i\) are the probabilities assigned by \(C_{\mathrm{det}}(\cdot)\) to the class ``fake''. Minimizing Eq.~(\ref{eq:l_rfdl}) encourages correct real/fake predictions on both clean and perturbed representations.

\noindent \textbf{Forensic Consistency Loss.}
The Forensic Consistency Loss (FCL) enforces consistent predictions for clean and perturbed representations, so that perturbation suppresses generator-specific shortcuts without altering the underlying forensic semantics. Let
\begin{equation}
p_i = p_{\mathrm{det}}(\cdot \mid \mathbf{H}_i), \qquad \tilde{p}_i = p_{\mathrm{det}}(\cdot \mid \tilde{\mathbf{H}}_i),
\label{eq:cons_prob}
\end{equation}
where \(p_{\mathrm{det}}(\cdot)\) is the output probability of the real/fake head. We define
\begin{equation}
\mathcal{L}_{\mathrm{FCL}}
=
\frac{1}{2N}
\sum_{i=1}^{N}
\left[
\mathrm{KL}\!\left(p_i \,\|\, \tilde{p}_i\right)
+
\mathrm{KL}\!\left(\tilde{p}_i \,\|\, p_i\right)
\right].
\label{eq:l_fcl}
\end{equation}
Minimizing Eq.~(\ref{eq:l_fcl}) preserves prediction consistency before and after perturbation, thereby discouraging reliance on unstable generator-dependent shortcuts.

\noindent \textbf{Total Loss.}
The overall objective for the main detection branch is
\begin{equation}
\mathcal{L}_{\mathrm{total}}
=
\mathcal{L}_{\mathrm{RFDL}}
+
\lambda_{\mathrm{FCL}}\mathcal{L}_{\mathrm{FCL}},
\label{eq:l_det_total}
\end{equation}
where \(\lambda_{\mathrm{FCL}}\) is a balancing coefficient. During training, the shared backbone and the real/fake head are optimized with \(\mathcal{L}_{\mathrm{total}}\), while the generator-ID head is updated separately with \(\mathcal{L}_{\mathrm{ID}}\).

\subsection{Adversarial Training}
\label{sec:adv}

We propose a token-aware adversarial training framework in which the attacker and detector are updated within the same training step to achieve co-evolution. Unlike conventional adversarial training methods that mainly improve robustness by maximizing classification difficulty under fixed perturbation budgets or by using static adversarial examples~\cite{wang2022deepfake,khan2024adversarially,li2025iron,li2026learning}, our framework is designed to suppress generator-specific shortcuts and improve open-world OOD generalization. In particular, the attacker in \methodname\ does not aim to directly fool the detector. Instead, it identifies and perturbs generator-sensitive yet detection-safe tokens in feature space, encouraging the detector to reduce reliance on generator-dependent cues and learn more transferable generator-invariant forensic traces. Since token importance is estimated with respect to the current detector, the adversarial signal evolves during training and remains aligned with the detector's changing decision boundary.

\noindent \textbf{Training Process.}
In each training step \(t\), we first update the attacker while freezing the detector. The scoring network is optimized with Eq.~(\ref{eq:l_att}) to fit the current refined token priority, and the updated attacker then selects the top-\(m\) important tokens to construct perturbed representations \(\tilde{\mathbf H}\) via gradient-guided embedding perturbation. Next, we freeze the attacker and update the detector: the generator-ID head is optimized with \(\mathcal{L}_{\mathrm{ID}}\), while the shared backbone and the real/fake head are optimized with Eq.~(\ref{eq:l_det_total}) on clean and perturbed representations. 
Through this alternating optimization, the attacker continuously exposes generator-specific shortcuts, while the detector progressively learns to preserve forensic semantics and become less dependent on generator-specific patterns.
We further use a linear ramp-up schedule for \(\alpha_t\), increasing it from 0 to \(\alpha_{\max}\) during the first \(T_{\mathrm{ramp}}\) steps, so that perturbation becomes progressively more cautious on detection-critical tokens. The complete procedure is provided as Algorithm~1 in the Appendix.

\newcommand{\gen}[1]{\textit{#1}}
\newcommand{\fdgen}[1]{\gen{#1}\textsuperscript{*}}
\newcommand{\macc}{\textit{Acc} $\uparrow$}
\newcommand{\mfone}{\textit{F1} $\uparrow$}
\newcommand{\maccd}{\textit{Acc} $\downarrow$}
\newcommand{\mfoned}{\textit{F1} $\downarrow$}

\section{Experiments and Results}

\subsection{Detection Performance in Open-world Settings}
\label{sec:ood_detection}

\noindent\textbf{Experiment Settings.}
We evaluate the open-world generalization of \methodname\ on \textbf{29 held-out generator-labeled entries} collected from VideoGen-RewardBench~\cite{videogenrewardbench_hf}, T2V-CompBench~\cite{t2vcompbench_hf}, and GenVidBench~\cite{genvidbench_hf}, covering both commercial models (e.g., Sora and Veo3) and open-source models (e.g., CogVideoX~\cite{yang2024cogvideox}, Open-Sora~\cite{opensora}, Wan2.6~\cite{wan2025wan}, and Latte~\cite{ma2024latte}). All methods share the same 10-generator ID training set. Real videos are sampled from Vript~\cite{vript_hf}, and all videos are uniformly sampled to 8 frames at \(224 \times 224\). For each evaluation entry, we initially pair generated videos with an equal number of real videos; reported metrics are computed on the videos retained after input-integrity checks, so a few final test sets are not exactly balanced. We use CLIP ViT-B/16~\cite{radford2021clip} as the default backbone and train \methodname\ for 12 epochs, with \(\epsilon = 0.3\), \(\xi = 0.01\), \(m = 10\% \times s\), and \(\lambda_{\mathrm{FCL}} = 0.1\). We compare against eight baselines: CLIP ViT-B/16~\cite{radford2021clip}, DINOv2-Base~\cite{oquab2023dinov2}, VideoMAEv2~\cite{wang2023videomae}, AIGVDet~\cite{bai2024ai}, DeMamba~\cite{chen2024demamba}, D3~\cite{zheng2025d3}, DeCoF~\cite{decof2025}, and UNITE~\cite{unite2025}. More details are provided in the Appendix.
The 29 entries are absent from ID training, although some share identifiable model-family lineages with ID generators. We therefore also evaluate an 18-entry subset that excludes these overlaps, using the same trained detectors, preprocessing, and thresholds; its construction is detailed in Appendix~\ref{sec:family_disjoint}.

\begin{table}[!t]
\centering
\setlength{\tabcolsep}{3pt}
\setlength{\arrayrulewidth}{0.3pt}
\setlength{\aboverulesep}{0.5pt}
\setlength{\belowrulesep}{0.5pt}
\renewcommand{\arraystretch}{0.82}
\scriptsize
\caption{Per-generator Acc/F1 on 29 held-out entries, with the 29-entry OOD and 18-entry family-disjoint means, ID mean, ID--OOD gap, and worst-case scores. \textsuperscript{*} marks entries in the 18-entry mean. Best in each row is \textbf{bold}; second-best is \underline{underlined}.}
\label{tab:per_generator_full}
\begin{adjustbox}{width=\textwidth}
\begin{tabular}{ccccccccccc}
\toprule
Generator & Metric & CLIP & DINO & VideoMAEv2 & AIGVDet & DeMamba & D3 & DeCoF & UNITE & \methodname \\
\midrule

\multirow{2}{*}{\gen{CogVideoX}} & \macc & \underline{95.29} & 86.14 & 71.14 & 70.57 & 90.83 & 67.57 & 82.43 & 85.57 & \textbf{98.86} \\
 & \mfone & \underline{95.13} & 84.58 & 61.15 & 58.30 & 89.92 & 68.52 & 79.60 & 83.48 & \textbf{98.86} \\
\hdashline[0.8pt/1.2pt]
\multirow{2}{*}{\fdgen{FastWan2.2}} & \macc & 90.62 & 90.18 & 76.00 & 55.86 & \underline{95.90} & 77.14 & 81.14 & 82.00 & \textbf{96.63} \\
 & \mfone & 89.37 & 88.96 & 70.00 & 21.77 & \underline{95.74} & 78.84 & 78.07 & 75.97 & \textbf{96.46} \\
\hdashline[0.8pt/1.2pt]
\multirow{2}{*}{\fdgen{HunyuanV-1.5}} & \macc & \underline{93.85} & 85.98 & 73.00 & 64.00 & 93.55 & 66.00 & 81.29 & 87.14 & \textbf{98.71} \\
 & \mfone & \underline{93.51} & 84.04 & 64.27 & 44.00 & 93.12 & 66.85 & 77.76 & 86.17 & \textbf{98.73} \\
\hdashline[0.8pt/1.2pt]
\multirow{2}{*}{\gen{Kelin2.6}} & \macc & \underline{97.99} & 95.98 & 68.50 & 63.33 & 95.30 & 67.17 & 91.33 & 91.33 & \textbf{98.49} \\
 & \mfone & \underline{97.97} & 95.96 & 55.53 & 42.11 & 95.08 & 69.17 & 90.78 & 91.53 & \textbf{98.50} \\
\hdashline[0.8pt/1.2pt]
\multirow{2}{*}{\fdgen{LTX-2}} & \macc & \underline{92.57} & 69.86 & 70.14 & 54.14 & 91.05 & 61.00 & 60.29 & 86.57 & \textbf{97.29} \\
 & \mfone & \underline{92.10} & 58.55 & 58.78 & 15.75 & 90.18 & 60.26 & 39.04 & 73.85 & \textbf{97.27} \\
\hdashline[0.8pt/1.2pt]
\multirow{2}{*}{\fdgen{Latte}} & \macc & 96.29 & 95.71 & 85.14 & 62.14 & \underline{98.84} & 73.00 & 93.57 & 96.43 & \textbf{98.86} \\
 & \mfone & 96.18 & 95.66 & 83.06 & 39.64 & \underline{98.84} & 75.23 & 93.43 & 96.29 & \textbf{98.86} \\
\hdashline[0.8pt/1.2pt]
\multirow{2}{*}{\fdgen{LongCat}} & \macc & \underline{94.00} & 90.71 & 75.57 & 70.71 & 93.21 & 72.86 & 87.00 & 89.00 & \textbf{96.43} \\
 & \mfone & \underline{93.67} & 89.95 & 69.30 & 58.59 & 92.73 & 75.39 & 86.02 & 87.98 & \textbf{96.31} \\
\hdashline[0.8pt/1.2pt]
\multirow{2}{*}{\gen{OpenSora-1.1}} & \macc & 94.43 & \underline{95.57} & 82.29 & 61.86 & 94.62 & 76.86 & 94.29 & 94.57 & \textbf{97.86} \\
 & \mfone & 94.17 & \underline{95.54} & 79.19 & 38.34 & 94.33 & 78.96 & 94.27 & 94.61 & \textbf{97.84} \\
\hdashline[0.8pt/1.2pt]
\multirow{2}{*}{\gen{OSP-v100}} & \macc & \underline{97.00} & 95.14 & 88.14 & 69.71 & 96.98 & 75.57 & 94.29 & 96.71 & \textbf{99.43} \\
 & \mfone & \underline{96.93} & 95.09 & 86.97 & 56.73 & 96.90 & 77.94 & 94.22 & 96.73 & \textbf{99.43} \\
\hdashline[0.8pt/1.2pt]
\multirow{2}{*}{\gen{OSP-v130}} & \macc & \underline{98.00} & 93.86 & 89.29 & 76.43 & 97.98 & 77.14 & 94.14 & 96.71 & \textbf{98.71} \\
 & \mfone & \underline{97.96} & 93.80 & 88.48 & 69.16 & 97.94 & 80.00 & 94.05 & 96.66 & \textbf{98.71} \\
\hdashline[0.8pt/1.2pt]
\multirow{2}{*}{\fdgen{OpenAI Sora}} & \macc & \underline{89.22} & 83.33 & 70.59 & 56.86 & 88.69 & 66.67 & \underline{89.22} & \textbf{91.18} & \textbf{91.18} \\
 & \mfone & 87.91 & 80.90 & 59.46 & 24.14 & 87.26 & 65.31 & \underline{89.11} & \textbf{90.53} & \textbf{90.53} \\
\hdashline[0.8pt/1.2pt]
\multirow{2}{*}{\gen{OpenSora}} & \macc & 80.00 & 84.71 & 59.43 & 58.57 & 81.15 & 81.00 & 85.71 & \underline{92.57} & \textbf{95.43} \\
 & \mfone & 75.35 & 82.77 & 35.75 & 29.61 & 76.79 & 83.19 & 84.08 & \underline{91.84} & \textbf{95.28} \\
\hdashline[0.8pt/1.2pt]
\multirow{2}{*}{\fdgen{Sora2}} & \macc & 68.05 & \underline{75.33} & 64.90 & 59.27 & 70.84 & 50.00 & \textbf{76.32} & 73.03 & 75.17 \\
 & \mfone & 54.16 & \underline{68.76} & 48.04 & 31.28 & 58.85 & 44.49 & \textbf{70.88} & 63.20 & 67.81 \\
\hdashline[0.8pt/1.2pt]
\multirow{2}{*}{\fdgen{Veo3}} & \macc & \underline{95.06} & 90.45 & 63.06 & 58.44 & 92.13 & 57.17 & 90.29 & 88.15 & \textbf{95.22} \\
 & \mfone & \underline{94.88} & 89.86 & 44.76 & 28.88 & 91.47 & 54.02 & 89.82 & 87.92 & \textbf{95.05} \\
\hdashline[0.8pt/1.2pt]
\multirow{2}{*}{\fdgen{Veo3-fast}} & \macc & \underline{95.60} & 90.09 & 62.58 & 57.70 & 94.32 & 55.50 & 90.72 & 88.01 & \textbf{96.86} \\
 & \mfone & \underline{95.39} & 89.19 & 44.13 & 26.70 & 93.99 & 52.12 & 90.12 & 87.63 & \textbf{96.75} \\
\hdashline[0.8pt/1.2pt]
\multirow{2}{*}{\fdgen{Wan2.6}} & \macc & \underline{88.17} & 84.50 & 74.33 & 53.00 & 86.05 & 57.83 & 82.00 & 85.83 & \textbf{91.17} \\
 & \mfone & \underline{86.78} & 82.49 & 66.52 & 11.32 & 83.80 & 56.60 & 78.74 & 81.60 & \textbf{90.55} \\
\hdashline[0.8pt/1.2pt]
\multirow{2}{*}{\fdgen{EasyAnimate}} & \macc & 81.00 & 77.71 & 70.57 & 64.43 & \underline{88.82} & 71.43 & 81.71 & 87.00 & \textbf{92.57} \\
 & \mfone & 76.87 & 72.63 & 60.08 & 44.79 & \underline{87.43} & 73.40 & 78.60 & 84.71 & \textbf{92.12} \\
\hdashline[0.8pt/1.2pt]
\multirow{2}{*}{\gen{Gen3}} & \macc & \underline{98.71} & 96.28 & 75.57 & 70.00 & 98.69 & 72.86 & 91.29 & 96.71 & \textbf{99.28} \\
 & \mfone & \underline{98.70} & 96.24 & 68.97 & 57.14 & 98.68 & 73.39 & 91.02 & 96.71 & \textbf{99.29} \\
\hdashline[0.8pt/1.2pt]
\multirow{2}{*}{\gen{Keling}} & \macc & 82.29 & \textbf{88.00} & 60.57 & 53.00 & 80.71 & 79.43 & \underline{86.57} & 81.86 & 85.71 \\
 & \mfone & 64.37 & 79.48 & 38.12 & 11.32 & 76.10 & \underline{81.63} & \textbf{85.76} & 73.78 & 73.87 \\
\hdashline[0.8pt/1.2pt]
\multirow{2}{*}{\gen{Kling}} & \macc & \underline{97.57} & 90.29 & 78.14 & 63.14 & 96.55 & 74.43 & 84.29 & 92.29 & \textbf{98.43} \\
 & \mfone & \underline{97.53} & 89.79 & 73.67 & 41.89 & 96.43 & 76.66 & 82.32 & 92.73 & \textbf{98.42} \\
\hdashline[0.8pt/1.2pt]
\multirow{2}{*}{\fdgen{Luma}} & \macc & \underline{97.71} & 89.86 & 67.86 & 57.86 & 97.67 & 68.57 & 83.43 & 93.14 & \textbf{97.86} \\
 & \mfone & \underline{97.67} & 89.09 & 55.97 & 27.16 & 97.62 & 69.70 & 81.11 & 93.62 & \textbf{97.85} \\
\hdashline[0.8pt/1.2pt]
\multirow{2}{*}{\fdgen{MiniMax}} & \macc & \textbf{95.71} & 93.29 & 79.14 & 58.29 & 91.51 & 72.71 & 79.14 & 93.43 & \underline{95.43} \\
 & \mfone & \textbf{95.59} & 93.08 & 74.48 & 28.43 & 90.74 & 75.10 & 74.74 & 93.12 & \underline{95.29} \\
\hdashline[0.8pt/1.2pt]
\multirow{2}{*}{\fdgen{MuSeV}} & \macc & 64.14 & 60.00 & 53.86 & 54.29 & 68.81 & \underline{72.00} & 55.43 & 66.00 & \textbf{79.43} \\
 & \mfone & 45.32 & 36.36 & 20.25 & 16.23 & 54.68 & \textbf{74.61} & 23.53 & 58.26 & \underline{74.29} \\
\hdashline[0.8pt/1.2pt]
\multirow{2}{*}{\gen{Pika}} & \macc & 93.13 & 85.41 & 79.71 & 74.57 & \underline{97.55} & 76.57 & 75.86 & 94.14 & \textbf{97.57} \\
 & \mfone & 92.75 & 83.65 & 76.01 & 66.03 & \underline{97.51} & 79.60 & 69.66 & 93.99 & \textbf{97.56} \\
\hdashline[0.8pt/1.2pt]
\multirow{2}{*}{\fdgen{Qingying}} & \macc & 62.43 & \underline{77.43} & 65.57 & 61.14 & 70.48 & 66.71 & 75.14 & \textbf{92.00} & 76.29 \\
 & \mfone & 40.36 & \underline{72.38} & 51.12 & 36.74 & 58.12 & 67.32 & 68.59 & \textbf{89.94} & 69.26 \\
\hdashline[0.8pt/1.2pt]
\multirow{2}{*}{\fdgen{T2VZ}} & \macc & \underline{92.86} & 90.43 & 89.43 & 52.71 & 91.54 & 74.71 & 85.71 & 91.86 & \textbf{97.14} \\
 & \mfone & \underline{92.40} & 89.86 & 88.51 & 10.78 & 90.77 & 76.49 & 83.92 & 91.33 & \textbf{97.08} \\
\hdashline[0.8pt/1.2pt]
\multirow{2}{*}{\fdgen{Tongyi}} & \macc & 82.00 & \textbf{94.86} & 73.71 & 65.43 & 78.49 & 73.71 & 88.86 & \underline{94.29} & 84.86 \\
 & \mfone & 78.28 & \textbf{94.77} & 66.30 & 47.16 & 72.60 & 76.29 & 88.22 & \underline{94.17} & 82.45 \\
\hdashline[0.8pt/1.2pt]
\multirow{2}{*}{\gen{VC2}} & \macc & \underline{97.71} & 95.14 & 91.29 & 70.71 & 97.69 & 79.43 & 91.14 & 96.14 & \textbf{99.00} \\
 & \mfone & \underline{97.68} & 95.11 & 90.88 & 58.92 & 97.65 & 82.40 & 90.83 & 96.01 & \textbf{99.00} \\
\hdashline[0.8pt/1.2pt]
\multirow{2}{*}{\fdgen{Vidu}} & \macc & 73.39 & 73.68 & 66.00 & 63.57 & 74.48 & 65.86 & 74.86 & \underline{88.71} & \textbf{92.99} \\
 & \mfone & 63.95 & 65.54 & 51.03 & 42.70 & 65.75 & 67.13 & 68.90 & \underline{87.67} & \textbf{92.54} \\
\midrule
\multirow{2}{*}{\textbf{OOD Mean (29)}} & \macc & 89.13 & 86.89 & 73.29 & 62.13 & \underline{89.46} & 70.03 & 83.71 & 89.39 & \textbf{93.89} \\
 & \mfone & 85.62 & 83.93 & 63.13 & 37.44 & 86.93 & 71.05 & 79.90 & \underline{87.31} & \textbf{92.62} \\
\midrule
\multirow{2}{*}{\textbf{Family-disjoint (18)}} & \macc & 86.26 & 84.08 & 71.19 & 59.44 & 87.02 & 66.83 & 80.90 & \underline{87.43} & \textbf{91.89} \\
 & \mfone & 81.91 & 80.11 & 59.78 & 30.89 & 83.54 & 67.18 & 75.59 & \underline{84.66} & \textbf{90.51} \\
\midrule
\multirow{2}{*}{\textbf{ID Mean}} & \macc & 98.15 & 97.25 & 91.10 & 76.95 & \underline{98.80} & 89.85 & 94.00 & 98.20 & \textbf{99.20} \\
 & \mfone & \underline{99.05} & 98.58 & 90.19 & 69.52 & 98.78 & 92.53 & 93.88 & 97.21 & \textbf{99.59} \\
\midrule
\multirow{2}{*}{\textbf{Gap (ID$-$OOD)}} & \maccd & 9.02 & 10.36 & 17.81 & 14.82 & 9.34 & 19.82 & 10.29 & \underline{8.81} & \textbf{5.31} \\
 & \mfoned & 13.43 & 14.65 & 27.06 & 32.08 & 11.85 & 21.48 & 13.98 & \underline{9.90} & \textbf{6.97} \\
\midrule
\multirow{2}{*}{\textbf{Worst-case}} & \macc & 62.43 & 60.00 & 53.86 & 52.71 & \underline{68.81} & 50.00 & 55.43 & 66.00 & \textbf{75.17} \\
 & \mfone & 40.36 & 36.36 & 20.25 & 10.78 & 54.68 & 44.49 & 23.53 & \underline{58.26} & \textbf{67.81} \\
\bottomrule
\end{tabular}
\end{adjustbox}
\end{table}

\begin{figure}[t]
    \centering
    \includegraphics[width=\textwidth]{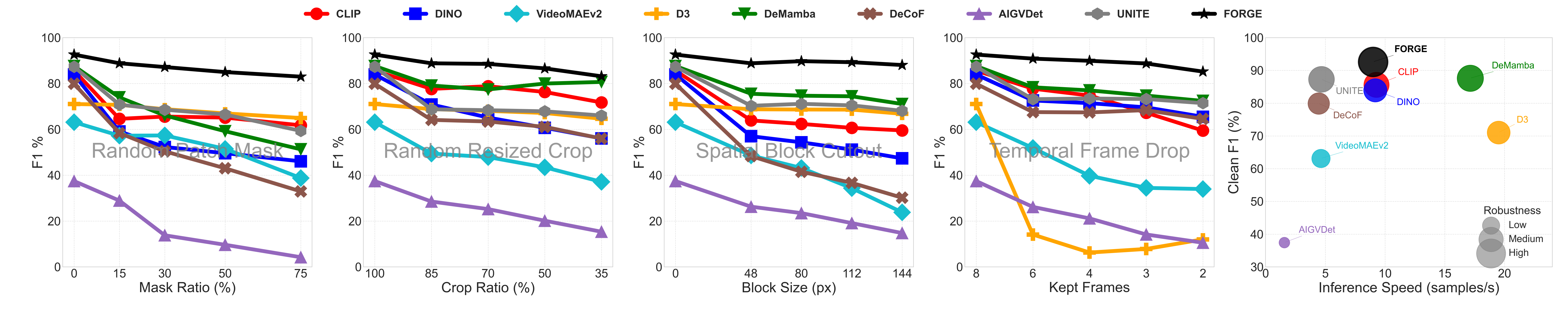}

    \caption{Detection performance under four post-processing perturbations (left) and the accuracy--throughput--robustness comparison (right). Higher mask ratio/block size and lower crop ratio/kept frames indicate stronger attacks; bubble size denotes mean F1 at the strongest level across all attacks.}
    \label{fig:robustness}

\end{figure}

\noindent\textbf{Experiment Results.} Table~\ref{tab:per_generator_full} reports per-generator results for the 29 held-out entries. \methodname\ achieves the highest mean Acc/F1 (\(93.89/92.62\)), exceeding the strongest baselines by \(4.43/5.31\) points. On the stricter 18-entry family-disjoint subset, \methodname\ reaches \(91.89/90.51\) Acc/F1 versus \(87.43/84.66\) for UNITE, retaining a \(5.85\)-point F1 lead (Table~\ref{tab:family_disjoint}). The advantage remains substantial after excluding entries with identifiable ID-family overlap, suggesting that the gain does not depend on those entries. On the broad benchmark, \methodname\ also attains the best worst-case Acc/F1 (\(75.17/67.81\)), exceeding the strongest competing scores by \(6.36/9.55\) points. Its ID--OOD gaps are the smallest at \(5.31/6.97\) points, \(3.50/2.93\) points below the next-best gaps. Together, these results support the value of reducing reliance on generator-specific shortcuts for transferable forensic detection.

\subsection{Robustness against Attacks}

\noindent\textbf{Experiment Setting.}
We test four post-processing perturbations at four strength levels on both real and fake videos: \textit{Random Patch Mask}, \textit{Spatial Block Cutout}, \textit{Random Resized Crop}, and \textit{Temporal Frame Drop}. These perturbations target local texture, spatial regions, geometric structure, and temporal continuity, respectively. We also measure inference throughput; Figure~\ref{fig:robustness} reports F1 across attack strengths and the accuracy--robustness--speed comparison.

\noindent\textbf{Experiment Results.}
Figure~\ref{fig:robustness} shows that \methodname\ achieves the best F1 curves under all four attacks and remains above 83\% at the strongest corruption levels. Its smaller degradation under local, spatial, geometric, and temporal perturbations supports the benefit of suppressing generator-specific shortcuts. Moreover, \methodname\ combines the highest clean F1 and largest robustness bubble with competitive throughput (8.99 samples/s), yielding the strongest overall accuracy--efficiency--robustness trade-off.

\subsection{Ablation Study}
\begin{wraptable}[15]{r}{0.43\textwidth}
\vspace{-\baselineskip}
\captionsetup{justification=raggedright,singlelinecheck=false}
\fontsize{7.5pt}{8.5pt}\selectfont
\setlength{\tabcolsep}{1.5pt}
\renewcommand{\arraystretch}{0.88}
\caption{F1 for the top-10 full-model entries; mean Acc/F1 over all 29. Bold/underlined: best/second-best.}
\label{tab:ablation}
\begin{adjustbox}{width=\linewidth}
\begin{tabular}{@{}lccccc@{}}
\toprule
Generator & Base & $-\mathrm{FCL}$ & $\alpha{=}0$ & $\alpha{=}.75$ & \methodname \\
\midrule
\textit{OSP-v100} & 96.93 & \underline{98.43} & 95.72 & 97.73 & \textbf{99.43} \\
\textit{Gen3} & \underline{98.70} & \textbf{99.29} & 98.55 & 98.16 & \textbf{99.29} \\
\textit{VC2} & 97.68 & 97.84 & \underline{98.12} & 97.01 & \textbf{99.00} \\
\textit{CogVideoX} & \underline{95.13} & 93.96 & 91.69 & 94.31 & \textbf{98.86} \\
\textit{Latte} & 96.18 & \underline{98.28} & 94.49 & 97.86 & \textbf{98.86} \\
\textit{HunyuanV-1.5} & 93.51 & 93.25 & 94.75 & \underline{94.99} & \textbf{98.73} \\
\textit{OSP-v130} & 97.96 & \underline{98.28} & 98.26 & 97.85 & \textbf{98.71} \\
\textit{Kelin2.6} & 97.97 & 98.33 & \textbf{98.64} & 98.33 & \underline{98.50} \\
\textit{Kling} & 97.53 & 97.70 & 96.47 & \underline{97.87} & \textbf{98.42} \\
\textit{Luma} & \underline{97.67} & \textbf{97.85} & 96.64 & 97.57 & \textbf{97.85} \\
\midrule
\textit{Avg. Acc} & 89.13 & 91.92 & 88.99 & \underline{92.37} & \textbf{93.89} \\
\textit{Avg. F1} & 85.62 & 90.10 & 85.49 & \underline{90.85} & \textbf{92.62} \\
\bottomrule
\end{tabular}
\end{adjustbox}
\end{wraptable}
\textbf{Ablation Setting.}
We compare the full \methodname\ with four variants: \textbf{Baseline} trains on clean videos without the attacker; \textbf{w/o \(\mathcal{L}_{\mathrm{FCL}}\)} removes consistency loss; \textbf{\(\alpha=0\)} selects tokens by generator relevance alone, omitting detection relevance in Eq.~(\ref{eq:s_final}); and \textbf{Fixed \(\alpha=0.75\)} replaces adaptive balancing with a fixed coefficient.

\textbf{Ablation Results.}
Table~\ref{tab:ablation} shows that the full \methodname\ reaches \(93.89\) accuracy and \(92.62\) F1. This improves on Fixed \(\alpha=0.75\) by \(1.52/1.77\) points and on Baseline by \(4.76/7.00\) points. Removing \(\mathcal{L}_{\mathrm{FCL}}\) lowers these scores to \(91.92/90.10\), confirming that consistency learning preserves real/fake evidence during perturbation. Using generator relevance alone (\(\alpha=0\)) further drops performance to \(88.99/85.49\), slightly below the baseline; thus effective token selection must jointly consider generator and detection relevance. Fixed weighting improves over Baseline but remains below the full model, indicating that adaptive balancing provides an additional benefit. Complete per-entry Acc/F1 results appear in Table~\ref{tab:ablation_full} of the Appendix.

\section{Discussion}

Additional analyses in the Appendix clarify the role of token selection. On held-out validation data, perturbing the top-scoring 10\% raises recall from \(0.79\) to \(0.89\), while random and low-score selection remain near baseline; the OOD test set is not used to choose this ratio. Replacing selected tokens lowers generator-ID accuracy from \(93.75\%\) to \(82.25\%\), with real/fake AUC changing only from \(99.94\%\) to \(99.50\%\). This pattern is consistent with selecting generator-informative tokens that are less critical to detection ranking.

Matched controls further narrow the explanation: Random-10\% and GRL reach \(86.03/87.34\) F1, versus \(92.62\) for \methodname, while gains also transfer to DINO and VideoMAE (about 20\% F1 for the latter). Under detector-aware PGD/FGSM, \methodname\ retains \(0.501/0.583\) F1 compared with CLIP's \(0.249/0.495\) and UNITE's \(0.182/0.213\). After matched H.264 re-encoding of five OOD entries, average AUC remains \(0.901/0.891/0.877\) across CRF18/23/28. Together, these results extend the evidence beyond the default backbone and common post-processing perturbations, although evaluation on longer videos and broader domains remains open.

\section{Conclusion}
We propose \methodname, a token-aware adversarial training framework for open-world AIGV detection. By selectively perturbing generator-sensitive yet detection-safe tokens, it reduces reliance on generator-specific shortcuts and encourages transferable, generator-invariant forensic representations.
Across 29 held-out benchmark entries, \methodname\ achieves 92.62 average F1, outperforming eight strong baselines by 5.31 points, while remaining robust under four common post-processing attacks. Ablations confirm the importance of forensic consistency learning and joint generator and detection relevance for token selection. Together, these results show the value of suppressing generator-dependent shortcuts for open-world AIGV detection.

\clearpage
\subsection*{AI use statement}
We used OpenAI Codex to improve the clarity of the manuscript, refine the presentation and discussion of existing experimental results, and adjust the LaTeX formatting. Generative AI was not used to design the method or experiments, generate or process research data, implement the method, or produce the reported experimental results. The authors take responsibility for the final content of this paper.

\bibliography{references}

\begin{thebibliography}{31}
\providecommand{\natexlab}[1]{#1}
\providecommand{\url}[1]{\texttt{#1}}
\expandafter\ifx\csname urlstyle\endcsname\relax
  \providecommand{\doi}[1]{doi: #1}\else
  \providecommand{\doi}{doi: \begingroup \urlstyle{rm}\Url}\fi

\bibitem[gen()]{genvidbench_hf}
Genvidbench.
\newblock \url{https://huggingface.co/datasets/jian-0/GenVidBench}.
\newblock Hugging Face repository, accessed 2026-03-29.

\bibitem[t2v()]{t2vcompbench_hf}
T2v-compbench-videos.
\newblock \url{https://huggingface.co/datasets/Kaiyue/T2V-CompBench-Videos}.
\newblock Hugging Face repository, accessed 2026-03-29.

\bibitem[vid()]{videogenrewardbench_hf}
Videogen-rewardbench.
\newblock \url{https://huggingface.co/datasets/KlingTeam/VideoGen-RewardBench}.
\newblock Hugging Face repository, accessed 2026-03-29.

\bibitem[vri()]{vript_hf}
Vript.
\newblock \url{https://huggingface.co/datasets/TIGER-Lab/Vript}.
\newblock Hugging Face dataset repository, accessed 2026-03-29.

\bibitem[Bai et~al.(2024)Bai, Lin, Cao, and Lou]{bai2024ai}
Jianfa Bai, Man Lin, Gang Cao, and Zijie Lou.
\newblock Ai-generated video detection via spatial-temporal anomaly learning.
\newblock In \emph{Chinese Conference on Pattern Recognition and Computer Vision (PRCV)}, pp.\  460--470. Springer, 2024.

\bibitem[Brooks et~al.(2024)Brooks, Peebles, Holmes, DePue, Guo, Jing, Schnurr, Taylor, Luhman, Luhman, et~al.]{brooks2024video}
Tim Brooks, Bill Peebles, Connor Holmes, Will DePue, Yufei Guo, Leo Jing, David Schnurr, Joe Taylor, Troy Luhman, Eric Luhman, et~al.
\newblock Video generation models as world simulators.
\newblock \emph{OpenAI Blog}, 1\penalty0 (8):\penalty0 1, 2024.

\bibitem[Chen et~al.(2024{\natexlab{a}})Chen, Zhang, Cun, Xia, Wang, Weng, and Shan]{chen2024videocrafter2}
Haoxin Chen, Yong Zhang, Xiaodong Cun, Menghan Xia, Xintao Wang, Chao Weng, and Ying Shan.
\newblock Videocrafter2: Overcoming data limitations for high-quality video diffusion models.
\newblock In \emph{Proceedings of the IEEE/CVF conference on computer vision and pattern recognition}, pp.\  7310--7320, 2024{\natexlab{a}}.

\bibitem[Chen et~al.(2024{\natexlab{b}})Chen, Hong, Huang, Xu, Gu, Li, Lan, Zhu, Zhang, Wang, et~al.]{chen2024demamba}
Haoxing Chen, Yan Hong, Zizheng Huang, Zhuoer Xu, Zhangxuan Gu, Yaohui Li, Jun Lan, Huijia Zhu, Jianfu Zhang, Weiqiang Wang, et~al.
\newblock Demamba: Ai-generated video detection on million-scale genvideo benchmark.
\newblock \emph{arXiv preprint arXiv:2405.19707}, 2024{\natexlab{b}}.

\bibitem[Collins \& Eck(2024)Collins and Eck]{collins2024new}
Eli Collins and Douglas Eck.
\newblock New generative media models and tools, built with and for creators, 2024.

\bibitem[Corvi et~al.(2025)Corvi, Cozzolino, Prashnani, De~Mello, Nagano, and Verdoliva]{corvi2025seeing}
Riccardo Corvi, Davide Cozzolino, Ekta Prashnani, Shalini De~Mello, Koki Nagano, and Luisa Verdoliva.
\newblock Seeing what matters: Generalizable ai-generated video detection with forensic-oriented augmentation.
\newblock \emph{arXiv preprint arXiv:2506.16802}, 2025.

\bibitem[Han et~al.(2025)Han, Huang, Hua, and Chen]{han2025towards}
Yue-Hua Han, Tai-Ming Huang, Kai-Lung Hua, and Jun-Cheng Chen.
\newblock Towards more general video-based deepfake detection through facial component guided adaptation for foundation model.
\newblock In \emph{Proceedings of the IEEE/CVF conference on computer vision and pattern recognition}, pp.\  22995--23005, 2025.

\bibitem[Intern{\`o} et~al.(2025)Intern{\`o}, Geirhos, Olhofer, Liu, Hammer, and Klindt]{interno2025ai}
Christian Intern{\`o}, Robert Geirhos, Markus Olhofer, Sunny Liu, Barbara Hammer, and David Klindt.
\newblock Ai-generated video detection via perceptual straightening.
\newblock \emph{arXiv preprint arXiv:2507.00583}, 2025.

\bibitem[Khan et~al.(2024)Khan, Chen, Liao, and Chen]{khan2024adversarially}
Sarwar Khan, Jun-Cheng Chen, Wen-Hung Liao, and Chu-Song Chen.
\newblock Adversarially robust deepfake detection via adversarial feature similarity learning.
\newblock In \emph{International conference on multimedia modeling}, pp.\  503--516. Springer, 2024.

\bibitem[Kundu et~al.(2025)Kundu, Xiong, Mohanty, Balachandran, and Roy-Chowdhury]{unite2025}
Rohit Kundu, Hao Xiong, Vishal Mohanty, Athula Balachandran, and Amit~K Roy-Chowdhury.
\newblock Towards a universal synthetic video detector: From face or background manipulations to fully ai-generated content.
\newblock In \emph{Proceedings of the Computer Vision and Pattern Recognition Conference}, pp.\  28050--28060, 2025.

\bibitem[Li et~al.(2025)Li, Zhang, Li, Shen, and Liu]{li2025iron}
Yuanfan Li, Zhaohan Zhang, Chengzhengxu Li, Chao Shen, and Xiaoming Liu.
\newblock Iron sharpens iron: Defending against attacks in machine-generated text detection with adversarial training.
\newblock In \emph{Proceedings of the 63rd Annual Meeting of the Association for Computational Linguistics (Volume 1: Long Papers)}, pp.\  3091--3113, 2025.

\bibitem[Li et~al.(2026)Li, Zhou, and Xie]{li2026learning}
Yuanfan Li, Qi~Zhou, and Zexuan Xie.
\newblock Learning from dictionary: Enhancing robustness of machine-generated text detection in zero-shot language via adversarial training.
\newblock In \emph{The Fourteenth International Conference on Learning Representations}, 2026.
\newblock URL \url{https://openreview.net/forum?id=bTcFHJo1Zk}.

\bibitem[Loshchilov \& Hutter(2017)Loshchilov and Hutter]{loshchilov2017decoupled}
Ilya Loshchilov and Frank Hutter.
\newblock Decoupled weight decay regularization.
\newblock \emph{arXiv preprint arXiv:1711.05101}, 2017.

\bibitem[Ma et~al.(2025)Ma, Yan, Guo, Liao, Yu, and Zhou]{decof2025}
Long Ma, Zhiyuan Yan, Qinglang Guo, Yong Liao, Haiyang Yu, and Pengyuan Zhou.
\newblock Detecting ai-generated video via frame consistency.
\newblock In \emph{2025 IEEE International Conference on Multimedia and Expo (ICME)}, pp.\  1--6. IEEE, 2025.

\bibitem[Ma et~al.(2024)Ma, Wang, Chen, Jia, Liu, Li, Chen, and Qiao]{ma2024latte}
Xin Ma, Yaohui Wang, Xinyuan Chen, Gengyun Jia, Ziwei Liu, Yuan-Fang Li, Cunjian Chen, and Yu~Qiao.
\newblock Latte: Latent diffusion transformer for video generation.
\newblock \emph{arXiv preprint arXiv:2401.03048}, 2024.

\bibitem[Oquab et~al.(2023)Oquab, Darcet, Moutakanni, Vo, Szafraniec, Khalidov, Fernandez, Haziza, Massa, El-Nouby, et~al.]{oquab2023dinov2}
Maxime Oquab, Timoth{\'e}e Darcet, Th{\'e}o Moutakanni, Huy Vo, Marc Szafraniec, Vasil Khalidov, Pierre Fernandez, Daniel Haziza, Francisco Massa, Alaaeldin El-Nouby, et~al.
\newblock Dinov2: Learning robust visual features without supervision.
\newblock \emph{arXiv preprint arXiv:2304.07193}, 2023.

\bibitem[Radford et~al.(2021)Radford, Kim, Hallacy, Ramesh, Goh, Agarwal, Sastry, Askell, Mishkin, Clark, et~al.]{radford2021clip}
Alec Radford, Jong~Wook Kim, Chris Hallacy, Aditya Ramesh, Gabriel Goh, Sandhini Agarwal, Girish Sastry, Amanda Askell, Pamela Mishkin, Jack Clark, et~al.
\newblock Learning transferable visual models from natural language supervision.
\newblock In \emph{International conference on machine learning}, pp.\  8748--8763. PmLR, 2021.

\bibitem[Selvaraju et~al.(2017)Selvaraju, Cogswell, Das, Vedantam, Parikh, and Batra]{selvaraju2017grad}
Ramprasaath~R Selvaraju, Michael Cogswell, Abhishek Das, Ramakrishna Vedantam, Devi Parikh, and Dhruv Batra.
\newblock Grad-cam: Visual explanations from deep networks via gradient-based localization.
\newblock In \emph{Proceedings of the IEEE international conference on computer vision}, pp.\  618--626, 2017.

\bibitem[Simonyan et~al.(2013)Simonyan, Vedaldi, and Zisserman]{simonyan2013deep}
Karen Simonyan, Andrea Vedaldi, and Andrew Zisserman.
\newblock Deep inside convolutional networks: Visualising image classification models and saliency maps.
\newblock \emph{arXiv preprint arXiv:1312.6034}, 2013.

\bibitem[Wan et~al.(2025)Wan, Wang, Ai, Wen, Mao, Xie, Chen, Yu, Zhao, Yang, et~al.]{wan2025wan}
Team Wan, Ang Wang, Baole Ai, Bin Wen, Chaojie Mao, Chen-Wei Xie, Di~Chen, Feiwu Yu, Haiming Zhao, Jianxiao Yang, et~al.
\newblock Wan: Open and advanced large-scale video generative models.
\newblock \emph{arXiv preprint arXiv:2503.20314}, 2025.

\bibitem[Wang et~al.(2023)Wang, Huang, Zhao, Tong, He, Wang, Wang, and Qiao]{wang2023videomae}
Limin Wang, Bingkun Huang, Zhiyu Zhao, Zhan Tong, Yinan He, Yi~Wang, Yali Wang, and Yu~Qiao.
\newblock Videomae v2: Scaling video masked autoencoders with dual masking.
\newblock In \emph{Proceedings of the IEEE/CVF conference on computer vision and pattern recognition}, pp.\  14549--14560, 2023.

\bibitem[Wang et~al.(2022)Wang, Guo, and Zuo]{wang2022deepfake}
Zhi Wang, Yiwen Guo, and Wangmeng Zuo.
\newblock Deepfake forensics via an adversarial game.
\newblock \emph{IEEE Transactions on Image Processing}, 31:\penalty0 3541--3552, 2022.

\bibitem[Yang et~al.(2024)Yang, Teng, Zheng, Ding, Huang, Xu, Yang, Hong, Zhang, Feng, et~al.]{yang2024cogvideox}
Zhuoyi Yang, Jiayan Teng, Wendi Zheng, Ming Ding, Shiyu Huang, Jiazheng Xu, Yuanming Yang, Wenyi Hong, Xiaohan Zhang, Guanyu Feng, et~al.
\newblock Cogvideox: Text-to-video diffusion models with an expert transformer.
\newblock \emph{arXiv preprint arXiv:2408.06072}, 2024.

\bibitem[Zhang et~al.(2025)Zhang, Lian, Yang, Li, Pang, Liu, Han, Li, and Tan]{zhang2025physics}
Shuhai Zhang, ZiHao Lian, Jiahao Yang, Daiyuan Li, Guoxuan Pang, Feng Liu, Bo~Han, Shutao Li, and Mingkui Tan.
\newblock Physics-driven spatiotemporal modeling for ai-generated video detection.
\newblock \emph{arXiv preprint arXiv:2510.08073}, 2025.

\bibitem[Zheng et~al.(2024{\natexlab{a}})Zheng, Lin, Zhao, Wang, Guo, Liu, and Shen]{zheng2024breaking}
Chende Zheng, Chenhao Lin, Zhengyu Zhao, Hang Wang, Xu~Guo, Shuai Liu, and Chao Shen.
\newblock Breaking semantic artifacts for generalized ai-generated image detection.
\newblock \emph{Advances in Neural Information Processing Systems}, 37:\penalty0 59570--59596, 2024{\natexlab{a}}.

\bibitem[Zheng et~al.(2025)Zheng, Suo, Lin, Zhao, Yang, Liu, Yang, Wang, and Shen]{zheng2025d3}
Chende Zheng, Ruiqi Suo, Chenhao Lin, Zhengyu Zhao, Le~Yang, Shuai Liu, Minghui Yang, Cong Wang, and Chao Shen.
\newblock D3: Training-free ai-generated video detection using second-order features.
\newblock In \emph{Proceedings of the IEEE/CVF International Conference on Computer Vision}, pp.\  12852--12862, 2025.

\bibitem[Zheng et~al.(2024{\natexlab{b}})Zheng, Peng, Yang, Shen, Li, Liu, Zhou, Li, and You]{opensora}
Zangwei Zheng, Xiangyu Peng, Tianji Yang, Chenhui Shen, Shenggui Li, Hongxin Liu, Yukun Zhou, Tianyi Li, and Yang You.
\newblock Open-sora: Democratizing efficient video production for all.
\newblock \emph{arXiv preprint arXiv:2412.20404}, 2024{\natexlab{b}}.

\end{thebibliography}
\bibliographystyle{iclr2027_conference}

\clearpage
\appendix
\raggedbottom
\section{Additional Experimental Details}

\subsection{Dataset}
\label{app:dataset}

We evaluate open-world AI-generated video detection using a fixed in-distribution (ID) / out-of-distribution (OOD) split. The ID set contains 10 generators: \textit{Gen-3}, \textit{Kling-1.0}, \textit{Pika-1.0}, \textit{PixVerse-V3}, \textit{Dreamina 1.2}, \textit{CogVideoX-5B}~\cite{yang2024cogvideox}, \textit{Open-Sora-1.2}~\cite{opensora}, \textit{Mochi}, \textit{VideoCrafter2}~\cite{chen2024videocrafter2}, and \textit{ZeroScope}. All compared methods are trained on this same 10-generator ID set.

For OOD evaluation, we use 29 held-out generator-labeled entries collected from three public benchmarks, \textit{VideoGen-RewardBench}~\cite{videogenrewardbench_hf}, \textit{T2V-CompBench}~\cite{t2vcompbench_hf}, and \textit{GenVidBench}~\cite{genvidbench_hf}. The 29 entries are: \textit{CogVideoX}, \textit{FastWan2.2}, \textit{HunyuanV-1.5}, \textit{Kelin2.6}, \textit{LTX-2}, \textit{Latte}~\cite{ma2024latte}, \textit{LongCat}, \textit{OpenSora-1.1}, \textit{OSP-v100}, \textit{OSP-v130}, \textit{OpenAI Sora}, \textit{OpenSora}, \textit{Sora2}, \textit{Veo3}, \textit{Veo3-fast}, \textit{Wan2.6}~\cite{wan2025wan}, \textit{EasyAnimate}, \textit{Gen3}, \textit{Keling}, \textit{Kling}, \textit{Luma}, \textit{MiniMax}, \textit{MuSeV}, \textit{Pika}, \textit{Qingying}, \textit{T2VZ}, \textit{Tongyi}, \textit{VC2}, and \textit{Vidu}. Here OOD denotes held-out benchmark entries; it does not imply generator-family disjointness for every entry. The ID CogVideoX-5B and OOD CogVideoX entries are drawn from different source collections and belong to the CogVideoX model family.

\noindent\textbf{Real-video source (Vript).}
All real videos are sampled from Vript~\cite{vript_hf}. For each generator-specific fake split, we initially sample an equal number of real videos from Vript. Input-integrity checks remove unreadable or invalid videos before scoring, which can leave unequal class counts: for example, the Keling split contains 175 generated and 350 real videos after 175 entries are removed at precheck. All reported Acc and F1 values use the retained samples. During OOD evaluation, the same real-video pool is shared across different generators to avoid introducing generator-dependent bias from real samples. All videos are uniformly sampled to 8 frames, resized to \(224 \times 224\), and processed with the same preprocessing pipeline.

\noindent\textbf{Generator sources by repository.}
The generator videos are collected from three public repositories, with some generator families appearing in more than one source. We retain the original repository-specific names here for reproducibility, while using normalized names in the main paper and tables.

\noindent\textbf{(1) Kaiyue/T2V-CompBench-Videos.}
This repository serves as the main source for the T2V-CompBench portion of our benchmark and contains several core ID generators together with widely used open-source baselines:
CogVideoX-5B, Dreamina\_1.2, Gen-2, Gen-3, Kling-1.0, Latte, Mochi, Open-Sora-1-1, Open-Sora-1-2, Open-Sora-Plan-v100, Open-Sora-Plan-v130, Pika-1.0, PixVerse-V3, T2V-Turbo-V2, Vico, VideoCrafter2, VideoTetris, and ZeroScope.

\noindent\textbf{(2) KlingTeam/VideoGen-RewardBench.}
This repository provides the broadest OOD coverage in our benchmark, including many recent commercial or frontier generators:
CogVideoX, easyanimatev4, FastWan2.2-TI2V-5B-FullAttn-Diffusers, Gen-3, HunyuanVideo-1.5, Kelin2\_6, Kling-1.0, kling1.5, LongCat-Video, LTX-2, luma, luma1.6, minimax, Open-Sora-1-2, qingying, Sora2, tongyi, Veo3\_1, Veo3\_1\_fast, vidu, and Wan2\_6.

\noindent\textbf{(3) jian-0/GenVidBench.}
This repository supplements the benchmark with additional challenging generators and complementary samples for several overlapping generator families:
Kelin2\_6, musev, OpenAI\_Sora, OpenSora\_13800, Pika-1.0, t2vz, and VideoCrafter2.

\subsection{Implementation}
\label{app:implementation}

We use CLIP ViT-B/16~\cite{radford2021clip} as the default backbone. Training runs for 12 epochs with an effective batch size of 32 under mixed precision (FP16). The backbone is fine-tuned with a learning rate of \(10^{-6}\), while all newly introduced modules, including the detection head, generator classifier, and scoring network, are optimized with a learning rate of \(10^{-4}\). We use AdamW~\cite{loshchilov2017decoupled} as the optimizer together with a cosine annealing schedule.

Adversarial perturbation is activated after a 2-epoch warmup, and its strength increases linearly during the early training stage. The perturbation hyperparameters are set to \(\epsilon = 0.3\), \(\xi = 0.01\), and \(m = 10\% * s\), meaning that the top \(10\%\) of tokens are selected for perturbation. The forensic consistency weight is set to \(\lambda_{\mathrm{FCL}} = 0.1\), and the scoring network is optimized directly with \(\mathcal{L}_{\mathrm{att}}\). The final decision threshold is selected on the validation set by maximizing the F1 score.

\subsection{Evaluation Metrics}
\label{app:metrics}

We report two standard binary classification metrics: accuracy (Acc) and F1 score (F1). Accuracy measures the overall proportion of correctly classified videos:
\begin{equation}
\mathrm{Acc} = \frac{TP + TN}{TP + TN + FP + FN},
\end{equation}
where \(TP\), \(TN\), \(FP\), and \(FN\) denote true positives, true negatives, false positives, and false negatives, respectively.

F1 is the harmonic mean of precision and recall:
\begin{equation}
\mathrm{F1} = \frac{2PR}{P + R},
\end{equation}
where
\begin{equation}
P = \frac{TP}{TP + FP}, \qquad
R = \frac{TP}{TP + FN}.
\end{equation}

In the main paper, we report both per-generator results and summary statistics across generators, including the OOD mean, ID mean, ID--OOD gap, and worst-case OOD performance.

\subsection{Experimental Comparison Methods}
\label{app:baselines}

To better evaluate the effectiveness of \methodname, we compare it with eight representative baselines spanning general visual encoders, video foundation models, and dedicated AI-generated video detectors.

\noindent\textbf{CLIP}~\cite{radford2021clip}: A large-scale vision--language pre-training model trained with contrastive learning on image--text pairs. CLIP learns transferable visual representations that have shown strong generalization across diverse downstream tasks. In our experiments, CLIP ViT-B/16 serves as a strong and widely used vision backbone baseline.

\noindent\textbf{DINOv2}~\cite{oquab2023dinov2}: A self-supervised visual representation learning method designed to produce robust and general-purpose image features without requiring labeled data. By scaling curated training data and model size, DINOv2 yields strong visual features that are effective across a broad range of image-level and pixel-level tasks. We use DINOv2-Base as a representative self-supervised visual baseline.

\noindent\textbf{VideoMAEv2}~\cite{wang2023videomae}: A scalable video masked autoencoder framework for learning video representations through self-supervised reconstruction. Its dual-masking design improves pre-training efficiency and supports strong transfer to downstream video understanding tasks. We include VideoMAEv2 as a representative video foundation model baseline.

\noindent\textbf{AIGVDet}~\cite{bai2024ai}: A dedicated AI-generated video detector based on spatio-temporal anomaly learning. It adopts a two-branch design that separately models spatial cues and optical-flow-based temporal anomalies, and then fuses both branches for final prediction. This method represents an early specialized detector for AIGV detection.

\noindent\textbf{DeMamba}~\cite{chen2024demamba}: A detector introduced together with the GenVideo benchmark for large-scale AI-generated video detection. It incorporates a plug-and-play \emph{Detail Mamba} module to capture local spatial--temporal inconsistencies and improve generalization and robustness under real-world degradations. We treat it as a strong recent supervised baseline.

\noindent\textbf{D3}~\cite{zheng2025d3}: A training-free AI-generated video detector based on second-order temporal features. Instead of relying on supervised fine-tuning, D3 exploits differences in second-order temporal dynamics between real and generated videos, making it a lightweight and efficient baseline for AIGV detection.

\noindent\textbf{DeCoF}~\cite{decof2025}: A detector based on frame consistency analysis. DeCoF emphasizes temporal consistency across frames while reducing the influence of static spatial shortcuts, thereby improving generalization to unseen video generators. It serves as a representative temporal-consistency-based AIGV detector.

\noindent\textbf{UNITE}~\cite{unite2025}: A universal synthetic video detector designed to handle both classical manipulated videos and fully AI-generated content. Unlike face-centric detectors, UNITE operates on full video frames and uses domain-agnostic features extracted from a strong vision foundation model, together with an attention-diversity objective to reduce over-reliance on facial regions. We include UNITE as a recent universal synthetic video detection baseline.

\subsection{Attack Methods}
\label{app:attacks}

We further evaluate robustness under four common post-processing perturbations: \textit{Random Patch Mask}, \textit{Spatial Block Cutout}, \textit{Random Resized Crop}, and \textit{Temporal Frame Drop}. These operations affect different types of forensic evidence and simulate distortions introduced during video editing or redistribution.

\noindent\textbf{Random Patch Mask.}
This attack randomly masks out multiple small local patches in individual frames. It mainly disrupts fine-grained local textures and weak forensic patterns that may be informative for distinguishing real and generated content.

\noindent\textbf{Spatial Block Cutout.}
This attack removes a larger contiguous spatial region from each frame. Compared with random patch masking, it introduces stronger spatial information loss and tests whether the detector remains reliable when larger semantic and structural regions are missing.

\noindent\textbf{Random Resized Crop.}
This attack first applies a random spatial crop and then resizes the cropped region back to the original resolution. It alters image composition, geometric structure, and local sampling statistics, thereby testing robustness to common re-framing and rescaling operations.

\noindent\textbf{Temporal Frame Drop.}
This attack randomly removes a subset of frames from the video and reconstructs the clip from the remaining frames. It primarily breaks temporal continuity and motion patterns, making it particularly useful for evaluating whether a detector relies on stable temporal forensic cues.

Each attack is evaluated at four progressively increasing strength levels and is applied to both real and fake videos. This protocol enables us to assess how performance degrades as post-processing distortion becomes more severe.

\section{Discussion}
\label{sec:appendix_discussion}

\begin{figure}[t]
    \centering
    \begin{subfigure}[b]{0.31\linewidth}
        \centering
        \includegraphics[width=\linewidth]{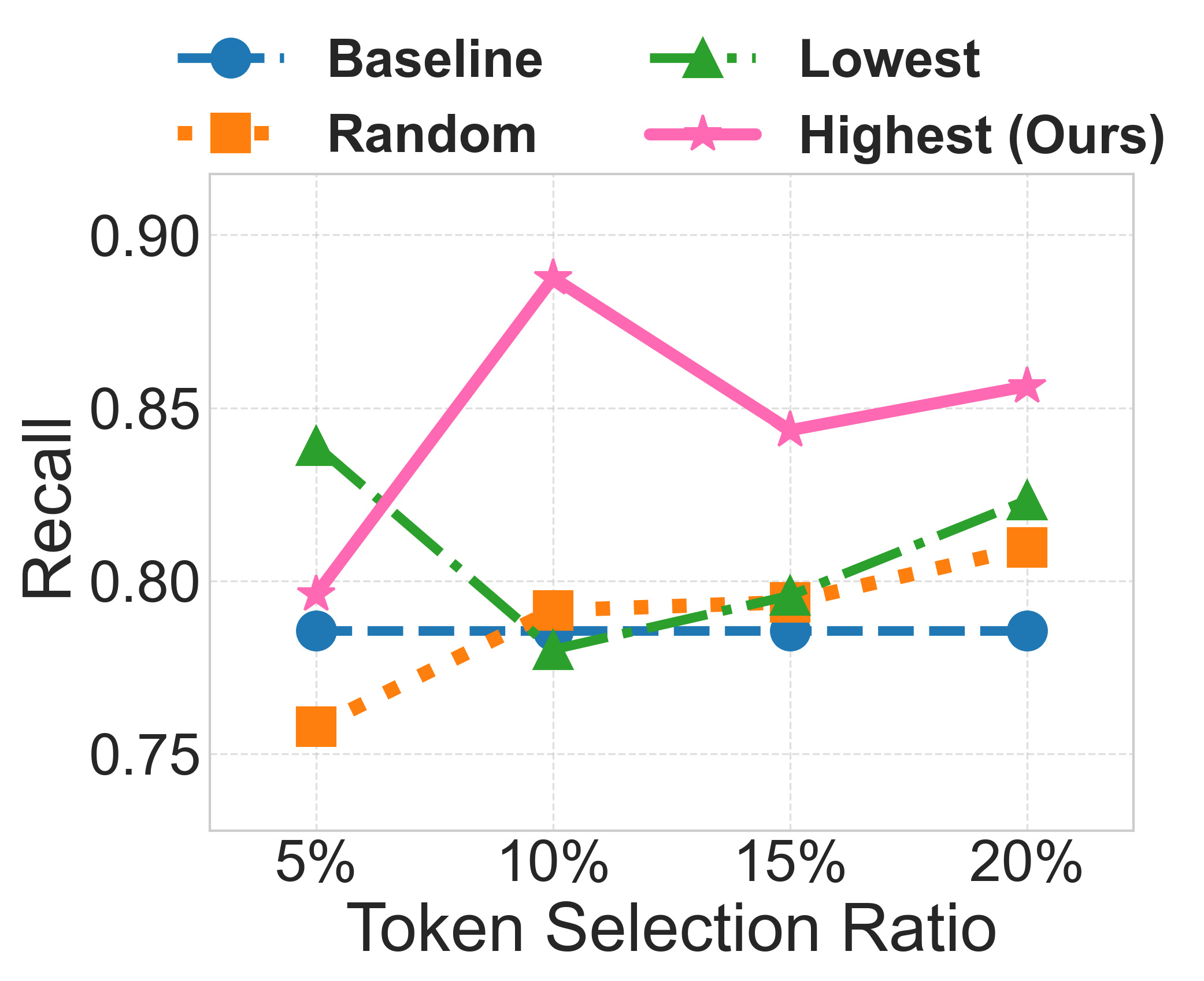}
        \caption{Effect of Token Ratio}
        \label{fig:token}
    \end{subfigure}%
    \hfill
    \begin{subfigure}[b]{0.31\linewidth}
        \centering
        \includegraphics[width=\linewidth]{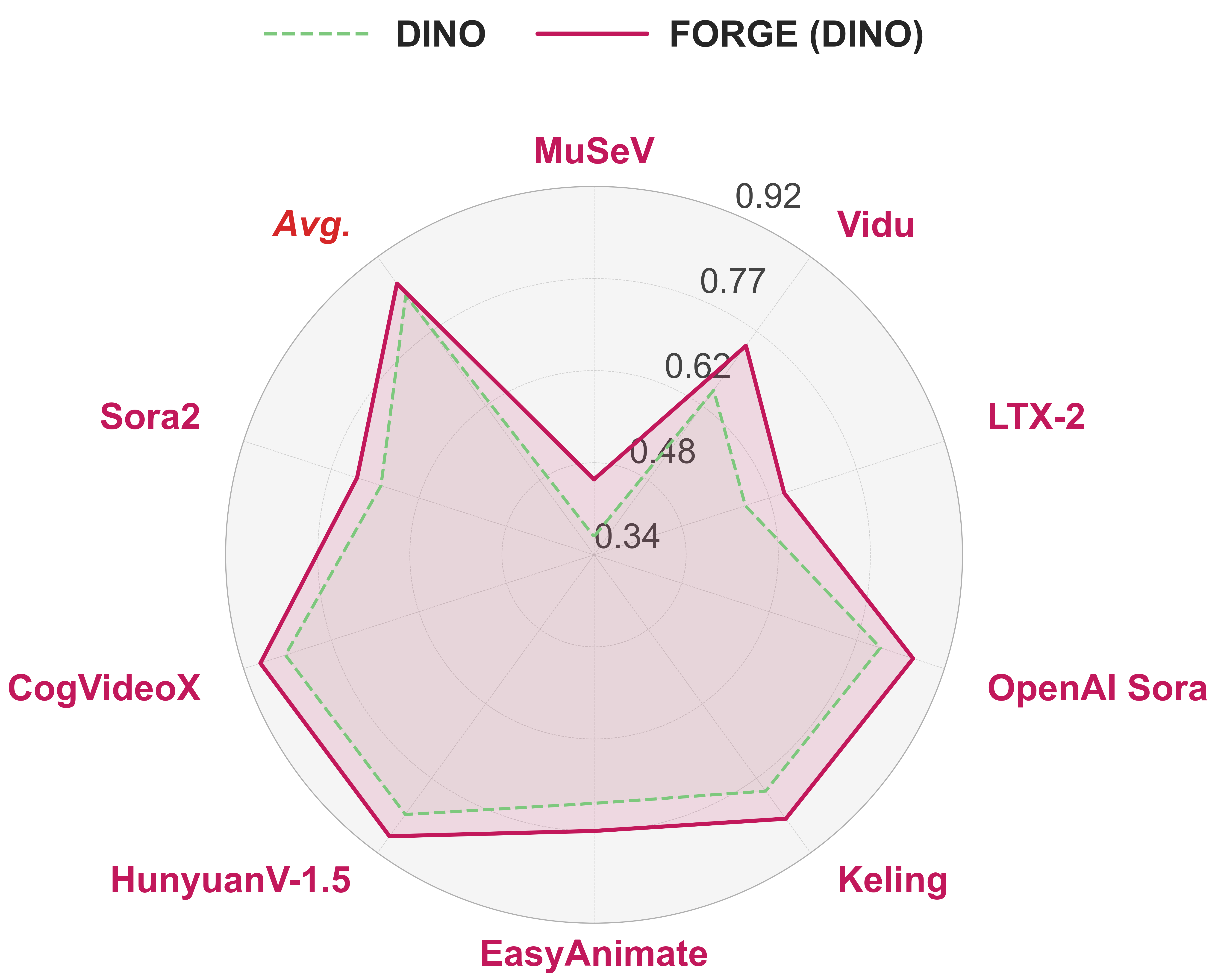}
        \caption{DINO Backbone}
        \label{fig:dino}
    \end{subfigure}%
    \hfill
    \begin{subfigure}[b]{0.31\linewidth}
        \centering
        \includegraphics[width=\linewidth]{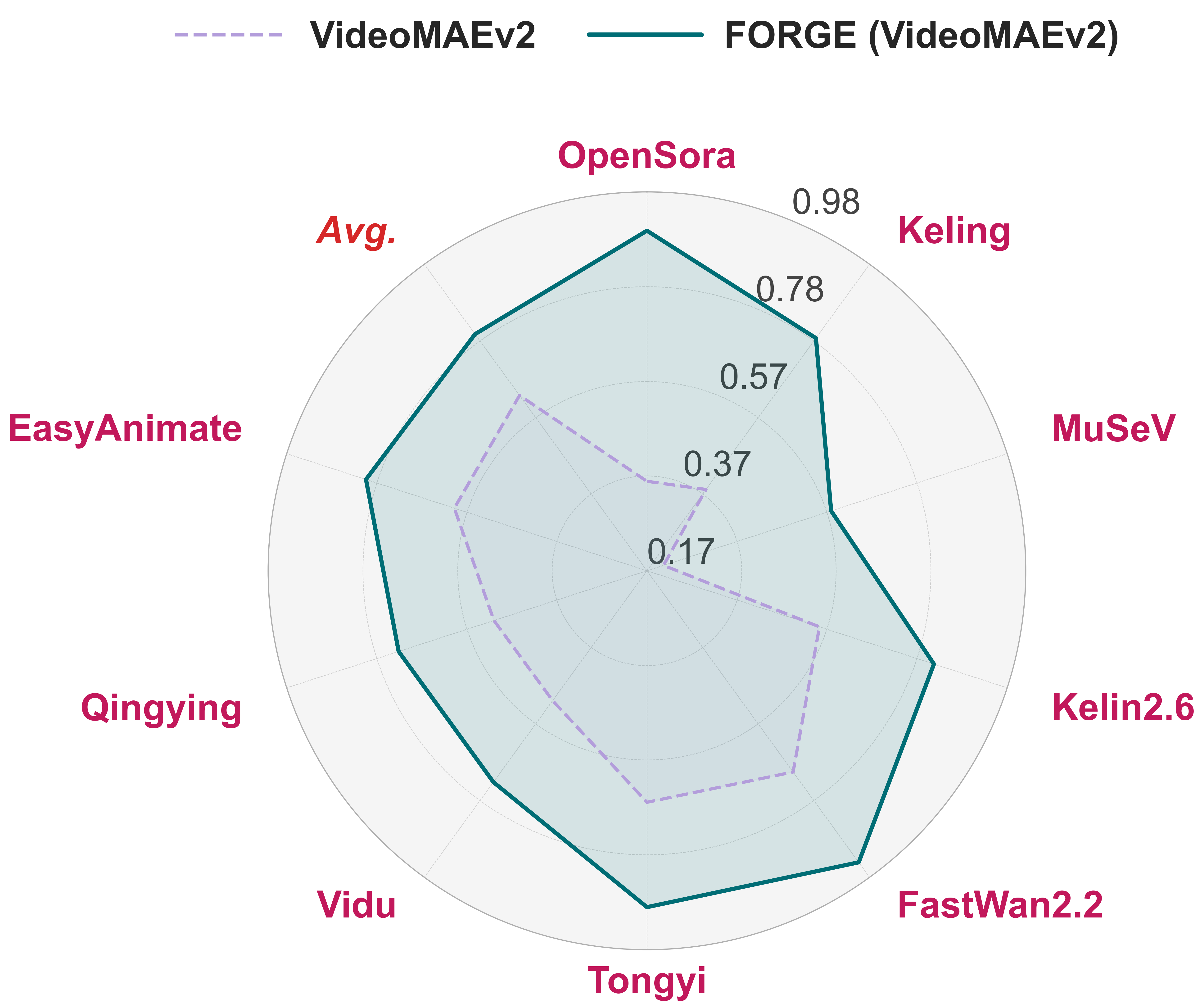}
        \caption{VideoMAE Backbone}
        \label{fig:videomae}
    \end{subfigure}
    \caption{\textbf{Additional validation and backbone analyses.}
    \textbf{(a)} Recall on a separate validation set under different token-selection ratios and perturbation strategies; the OOD test set is not used for this analysis or for hyperparameter selection.
    \textbf{(b)} F1 score of \methodname\ with DINO as the backbone.
    \textbf{(c)} F1 score of \methodname\ with VideoMAE as the backbone.}
    \label{fig:discussion}

\end{figure}

\subsection{Effect of Token Ratio and Selection Strategy}
\noindent\textbf{Experiment Setting.}
Figure~\ref{fig:token} reports a sensitivity analysis conducted exclusively on an additional validation set, rather than on the OOD test set. No test examples or test results are used to choose the token ratio or tune any hyperparameter. We vary the ratio of perturbed tokens and compare the highest-scoring, lowest-scoring, and random tokens while keeping the training protocol fixed.

\noindent\textbf{Experiment Result and Analysis.}
Random and low-score perturbation remain close to the baseline across all ratios, showing that perturbation alone provides little benefit. In contrast, perturbing the highest-scoring tokens raises validation recall from approximately \(0.79\) to \(0.89\), with the strongest result at 10\%. We therefore use 10\% as the validation-selected operating point without consulting OOD test performance. This comparison supports selective shortcut suppression and the need to preserve detection-critical evidence.

\subsection{Generality Across Backbones}
\noindent\textbf{Experiment Setting.}
Figures~\ref{fig:dino} and~\ref{fig:videomae} replace the default CLIP encoder with DINO and VideoMAE, respectively, while retaining the same OOD protocol and comparing each \methodname\ model with its matched backbone baseline.

\noindent\textbf{Experiment Result and Analysis.}
\methodname\ consistently improves OOD performance with both encoders, and the VideoMAE F1 gain is approximately 20\%. Thus, the benefit is not tied to a specific visual encoder or pretraining paradigm; suppressing generator-specific shortcuts yields more transferable forensic representations across backbone families.

\subsection{Family-Disjoint OOD Generalization}
\label{sec:family_disjoint}
\noindent\textbf{Subset construction.}
The main benchmark contains 29 evaluation entries absent from the training split, but some entries belong to a generator family also represented in ID training. In particular, ID CogVideoX-5B and OOD CogVideoX come from different source collections but share the CogVideoX family name. To assess transfer beyond these lineages, we conservatively screen for ID-family overlap using the original entry labels in the public benchmark repositories and associated model/project names. We remove the following 11 original benchmark entries: \texttt{CogVideoX}, \texttt{Kelin2\_6}, \texttt{Open-Sora-1-1}, \texttt{Open-Sora-Plan-v100}, \texttt{Open-Sora-Plan-v130}, \texttt{OpenSora\_13800}, \texttt{gen3}, \texttt{keling}, \texttt{kling}, \texttt{pika}, and \texttt{vc2}. These removals cover the CogVideoX, Open-Sora/OSP, Gen-3, Kling/Keling, Pika, and VideoCrafter/VC2 lineages represented in the ID training set. We keep the other 18 entries, for which this name-and-lineage screen does not identify an ID counterpart. In particular, the OpenAI Sora and Sora2 entries are retained: sharing the word ``Sora'' with Open-Sora is not by itself treated as evidence of common model lineage. This subset provides a stricter generator-family comparison within the existing benchmark; it does not remove the separate possibility of source-dataset differences between real and generated videos. No model is retrained, and the evaluation thresholds and preprocessing are unchanged.

\noindent\textbf{Results.}
Across the 18 retained generators, \methodname\ obtains 91.89 Acc and 90.51 F1. The strongest comparison in this subset is UNITE at 87.43 Acc and 84.66 F1, followed by the other baselines reported in Table~\ref{tab:family_disjoint}. Thus the mean advantage persists under this stricter split, although per-generator results remain heterogeneous. The table reuses the corresponding rows of the main benchmark and recomputes only the subset means.

\begin{table}[H]
\centering
\setlength{\tabcolsep}{3pt}
\renewcommand{\arraystretch}{0.86}
\scriptsize
\caption{Per-generator accuracy (Acc) and F1 on the 18-entry family-disjoint OOD subset. The mean is the unweighted average across the retained entries. Best in each row is \textbf{bold}; second-best is \underline{underlined}.}
\label{tab:family_disjoint}
\begin{adjustbox}{width=\textwidth}
\begin{tabular}{ccccccccccc}
\toprule
Generator & Metric & CLIP & DINO & VideoMAEv2 & AIGVDet & DeMamba & D3 & DeCoF & UNITE & \methodname \\
\midrule
\multirow{2}{*}{\textit{FastWan2.2}} & \textit{Acc} & 90.62 & 90.18 & 76.00 & 55.86 & \underline{95.90} & 77.14 & 81.14 & 82.00 & \textbf{96.63} \\
 & \textit{F1} & 89.37 & 88.96 & 70.00 & 21.77 & \underline{95.74} & 78.84 & 78.07 & 75.97 & \textbf{96.46} \\
\addlinespace[1pt]
\multirow{2}{*}{\textit{HunyuanV-1.5}} & \textit{Acc} & \underline{93.85} & 85.98 & 73.00 & 64.00 & 93.55 & 66.00 & 81.29 & 87.14 & \textbf{98.71} \\
 & \textit{F1} & \underline{93.51} & 84.04 & 64.27 & 44.00 & 93.12 & 66.85 & 77.76 & 86.17 & \textbf{98.73} \\
\addlinespace[1pt]
\multirow{2}{*}{\textit{LTX-2}} & \textit{Acc} & \underline{92.57} & 69.86 & 70.14 & 54.14 & 91.05 & 61.00 & 60.29 & 86.57 & \textbf{97.29} \\
 & \textit{F1} & \underline{92.10} & 58.55 & 58.78 & 15.75 & 90.18 & 60.26 & 39.04 & 73.85 & \textbf{97.27} \\
\addlinespace[1pt]
\multirow{2}{*}{\textit{Latte}} & \textit{Acc} & 96.29 & 95.71 & 85.14 & 62.14 & \underline{98.84} & 73.00 & 93.57 & 96.43 & \textbf{98.86} \\
 & \textit{F1} & 96.18 & 95.66 & 83.06 & 39.64 & \underline{98.84} & 75.23 & 93.43 & 96.29 & \textbf{98.86} \\
\addlinespace[1pt]
\multirow{2}{*}{\textit{LongCat}} & \textit{Acc} & \underline{94.00} & 90.71 & 75.57 & 70.71 & 93.21 & 72.86 & 87.00 & 89.00 & \textbf{96.43} \\
 & \textit{F1} & \underline{93.67} & 89.95 & 69.30 & 58.59 & 92.73 & 75.39 & 86.02 & 87.98 & \textbf{96.31} \\
\addlinespace[1pt]
\multirow{2}{*}{\textit{OpenAI Sora}} & \textit{Acc} & \underline{89.22} & 83.33 & 70.59 & 56.86 & 88.69 & 66.67 & \underline{89.22} & \textbf{91.18} & \textbf{91.18} \\
 & \textit{F1} & 87.91 & 80.90 & 59.46 & 24.14 & 87.26 & 65.31 & \underline{89.11} & \textbf{90.53} & \textbf{90.53} \\
\addlinespace[1pt]
\multirow{2}{*}{\textit{Sora2}} & \textit{Acc} & 68.05 & \underline{75.33} & 64.90 & 59.27 & 70.84 & 50.00 & \textbf{76.32} & 73.03 & 75.17 \\
 & \textit{F1} & 54.16 & \underline{68.76} & 48.04 & 31.28 & 58.85 & 44.49 & \textbf{70.88} & 63.20 & 67.81 \\
\addlinespace[1pt]
\multirow{2}{*}{\textit{Veo3}} & \textit{Acc} & \underline{95.06} & 90.45 & 63.06 & 58.44 & 92.13 & 57.17 & 90.29 & 88.15 & \textbf{95.22} \\
 & \textit{F1} & \underline{94.88} & 89.86 & 44.76 & 28.88 & 91.47 & 54.02 & 89.82 & 87.92 & \textbf{95.05} \\
\addlinespace[1pt]
\multirow{2}{*}{\textit{Veo3-fast}} & \textit{Acc} & \underline{95.60} & 90.09 & 62.58 & 57.70 & 94.32 & 55.50 & 90.72 & 88.01 & \textbf{96.86} \\
 & \textit{F1} & \underline{95.39} & 89.19 & 44.13 & 26.70 & 93.99 & 52.12 & 90.12 & 87.63 & \textbf{96.75} \\
\addlinespace[1pt]
\multirow{2}{*}{\textit{Wan2.6}} & \textit{Acc} & \underline{88.17} & 84.50 & 74.33 & 53.00 & 86.05 & 57.83 & 82.00 & 85.83 & \textbf{91.17} \\
 & \textit{F1} & \underline{86.78} & 82.49 & 66.52 & 11.32 & 83.80 & 56.60 & 78.74 & 81.60 & \textbf{90.55} \\
\addlinespace[1pt]
\multirow{2}{*}{\textit{EasyAnimate}} & \textit{Acc} & 81.00 & 77.71 & 70.57 & 64.43 & \underline{88.82} & 71.43 & 81.71 & 87.00 & \textbf{92.57} \\
 & \textit{F1} & 76.87 & 72.63 & 60.08 & 44.79 & \underline{87.43} & 73.40 & 78.60 & 84.71 & \textbf{92.12} \\
\addlinespace[1pt]
\multirow{2}{*}{\textit{Luma}} & \textit{Acc} & \underline{97.71} & 89.86 & 67.86 & 57.86 & 97.67 & 68.57 & 83.43 & 93.14 & \textbf{97.86} \\
 & \textit{F1} & \underline{97.67} & 89.09 & 55.97 & 27.16 & 97.62 & 69.70 & 81.11 & 93.62 & \textbf{97.85} \\
\addlinespace[1pt]
\multirow{2}{*}{\textit{MiniMax}} & \textit{Acc} & \textbf{95.71} & 93.29 & 79.14 & 58.29 & 91.51 & 72.71 & 79.14 & 93.43 & \underline{95.43} \\
 & \textit{F1} & \textbf{95.59} & 93.08 & 74.48 & 28.43 & 90.74 & 75.10 & 74.74 & 93.12 & \underline{95.29} \\
\addlinespace[1pt]
\multirow{2}{*}{\textit{MuSeV}} & \textit{Acc} & 64.14 & 60.00 & 53.86 & 54.29 & 68.81 & \underline{72.00} & 55.43 & 66.00 & \textbf{79.43} \\
 & \textit{F1} & 45.32 & 36.36 & 20.25 & 16.23 & 54.68 & \textbf{74.61} & 23.53 & 58.26 & \underline{74.29} \\
\addlinespace[1pt]
\multirow{2}{*}{\textit{Qingying}} & \textit{Acc} & 62.43 & \underline{77.43} & 65.57 & 61.14 & 70.48 & 66.71 & 75.14 & \textbf{92.00} & 76.29 \\
 & \textit{F1} & 40.36 & \underline{72.38} & 51.12 & 36.74 & 58.12 & 67.32 & 68.59 & \textbf{89.94} & 69.26 \\
\addlinespace[1pt]
\multirow{2}{*}{\textit{T2VZ}} & \textit{Acc} & \underline{92.86} & 90.43 & 89.43 & 52.71 & 91.54 & 74.71 & 85.71 & 91.86 & \textbf{97.14} \\
 & \textit{F1} & \underline{92.40} & 89.86 & 88.51 & 10.78 & 90.77 & 76.49 & 83.92 & 91.33 & \textbf{97.08} \\
\addlinespace[1pt]
\multirow{2}{*}{\textit{Tongyi}} & \textit{Acc} & 82.00 & \textbf{94.86} & 73.71 & 65.43 & 78.49 & 73.71 & 88.86 & \underline{94.29} & 84.86 \\
 & \textit{F1} & 78.28 & \textbf{94.77} & 66.30 & 47.16 & 72.60 & 76.29 & 88.22 & \underline{94.17} & 82.45 \\
\addlinespace[1pt]
\multirow{2}{*}{\textit{Vidu}} & \textit{Acc} & 73.39 & 73.68 & 66.00 & 63.57 & 74.48 & 65.86 & 74.86 & \underline{88.71} & \textbf{92.99} \\
 & \textit{F1} & 63.95 & 65.54 & 51.03 & 42.70 & 65.75 & 67.13 & 68.90 & \underline{87.67} & \textbf{92.54} \\
\midrule
\multirow{2}{*}{\textbf{OOD Mean}} & \textit{Acc} & 86.26 & 84.08 & 71.19 & 59.44 & 87.02 & 66.83 & 80.90 & \underline{87.43} & \textbf{91.89} \\
 & \textit{F1} & 81.91 & 80.11 & 59.78 & 30.89 & 83.54 & 67.18 & 75.59 & \underline{84.66} & \textbf{90.51} \\
\bottomrule
\end{tabular}
\end{adjustbox}
\end{table}

\subsection{Complete Ablation Results}
Table~\ref{tab:ablation_full} retains both accuracy and F1 for each of the ten generators summarized in the compact main-paper table. The ten displayed generators have the highest F1 under the full model; the OOD mean is computed over all 29 evaluation entries.

\begin{table}[H]
\centering
\footnotesize
\setlength{\tabcolsep}{3pt}
\setlength{\arrayrulewidth}{0.3pt}
\setlength{\aboverulesep}{0.3pt}
\setlength{\belowrulesep}{0.3pt}
\renewcommand{\arraystretch}{0.85}

\caption{Full Acc and F1 ablation results for the 10 generators displayed in the main-paper ablation table; the OOD mean covers all 29 evaluation entries. Best results are \textbf{bold}; second-best results are \underline{underlined}.}
\label{tab:ablation_full}
\begin{adjustbox}{max width=\textwidth}
\begin{tabular}{llccccc}
\toprule
Generator & Metric & Baseline & \makecell{w/o \\ $\mathcal{L}_{\mathrm{FCL}}$} & \makecell{w/o balancing \\ coefficient ($\alpha=0$)} & \makecell{Fixed \\ $\alpha=0.75$} & \methodname \\
\midrule
\multirow{2}{*}{\textit{OSP-v100}} 
& \textit{Acc} $\uparrow$ & 97.00 & \underline{98.43} & 95.86 & 97.71 & \textbf{99.43} \\
& \textit{F1} $\uparrow$  & 96.93 & \underline{98.43} & 95.72 & 97.73 & \textbf{99.43} \\
\addlinespace[1pt]
\multirow{2}{*}{\textit{Gen3}} 
& \textit{Acc} $\uparrow$ & 98.71 & \underline{99.28} & 98.57 & 98.14 & \textbf{99.28} \\
& \textit{F1} $\uparrow$  & 98.70 & \underline{99.29} & 98.55 & 98.16 & \textbf{99.29} \\
\addlinespace[1pt]
\multirow{2}{*}{\textit{VC2}} 
& \textit{Acc} $\uparrow$ & 97.71 & 97.86 & \underline{98.14} & 97.00 & \textbf{99.00} \\
& \textit{F1} $\uparrow$  & 97.68 & 97.84 & \underline{98.12} & 97.01 & \textbf{99.00} \\
\addlinespace[1pt]
\multirow{2}{*}{\textit{CogVideoX}} 
& \textit{Acc} $\uparrow$ & \underline{95.29} & 94.14 & 92.29 & 94.43 & \textbf{98.86} \\
& \textit{F1} $\uparrow$  & \underline{95.13} & 93.96 & 91.69 & 94.31 & \textbf{98.86} \\
\addlinespace[1pt]
\multirow{2}{*}{\textit{Latte}} 
& \textit{Acc} $\uparrow$ & 96.29 & \underline{98.29} & 94.71 & 97.86 & \textbf{98.86} \\
& \textit{F1} $\uparrow$  & 96.18 & \underline{98.28} & 94.49 & 97.86 & \textbf{98.86} \\
\addlinespace[1pt]
\multirow{2}{*}{\textit{HunyuanV-1.5}} 
& \textit{Acc} $\uparrow$ & 93.85 & 93.56 & 94.99 & \underline{95.14} & \textbf{98.71} \\
& \textit{F1} $\uparrow$  & 93.51 & 93.25 & 94.75 & \underline{94.99} & \textbf{98.73} \\
\addlinespace[1pt]
\multirow{2}{*}{\textit{OSP-v130}} 
& \textit{Acc} $\uparrow$ & 98.00 & \underline{98.29} & \underline{98.29} & 97.86 & \textbf{98.71} \\
& \textit{F1} $\uparrow$  & 97.96 & \underline{98.28} & 98.26 & 97.85 & \textbf{98.71} \\
\addlinespace[1pt]
\multirow{2}{*}{\textit{Kelin2.6}} 
& \textit{Acc} $\uparrow$ & 97.99 & 98.32 & \textbf{98.66} & 98.32 & \underline{98.49} \\
& \textit{F1} $\uparrow$  & 97.97 & 98.33 & \textbf{98.64} & 98.33 & \underline{98.50} \\
\addlinespace[1pt]
\multirow{2}{*}{\textit{Kling}} 
& \textit{Acc} $\uparrow$ & 97.57 & 97.71 & 96.57 & \underline{97.86} & \textbf{98.43} \\
& \textit{F1} $\uparrow$  & 97.53 & 97.70 & 96.47 & \underline{97.87} & \textbf{98.42} \\
\addlinespace[1pt]
\multirow{2}{*}{\textit{Luma}} 
& \textit{Acc} $\uparrow$ & 97.71 & \underline{97.86} & 96.71 & 97.57 & \textbf{97.86} \\
& \textit{F1} $\uparrow$  & 97.67 & \underline{97.85} & 96.64 & 97.57 & \textbf{97.85} \\
\midrule

\multirow{2}{*}{OOD Mean} 
& \textit{Acc} $\uparrow$ & 89.13 & 91.92 & 88.99 & \underline{92.37} & \textbf{93.89} \\
& \textit{F1} $\uparrow$  & 85.62 & 90.10 & 85.49 & \underline{90.85} & \textbf{92.62} \\
\bottomrule
\end{tabular}
\end{adjustbox}

\end{table}

\subsection{Mechanism Isolation}
\noindent\textbf{Experiment Setting.}
We compare two matched controls with the baseline and full model. Random-10\% replaces score-based selection with uniformly sampled tokens while keeping the same perturbation ratio, strength, and FCL weight \(\lambda_{\mathrm{FCL}}=0.1\); it does not use the scoring loss. GRL applies gradient reversal to encourage globally generator-invariant features. All variants use the same ID training set, backbone, and evaluation over the 29 held-out entries.

\begin{table}[H]
\centering
\scriptsize
\setlength{\tabcolsep}{4pt}
\renewcommand{\arraystretch}{0.9}
\begin{tabular}{lccc}
\toprule
Variant & Control purpose & F1 & Acc \\
\midrule
Baseline & no adversarial perturbation & 85.62 & 89.13 \\
Random-10\% & generic token perturbation & 86.03 & 89.34 \\
GRL & global generator invariance & 87.34 & 89.71 \\
\methodname & selective suppression & \textbf{92.62} & \textbf{93.89} \\
\bottomrule
\end{tabular}
\caption{Mechanism-isolation results on the 29 held-out entries.}
\label{tab:mechanism_isolation}

\end{table}

\noindent\textbf{Experiment Result and Analysis.}
Table~\ref{tab:mechanism_isolation} shows only marginal improvements from Random-10\% and GRL, whereas \methodname\ improves baseline F1 by 7.00 points. Together with the main-paper \(\alpha=0\) ablation, this result shows that neither generic noise nor global generator confusion explains the gain; it depends on selectively perturbing generator-sensitive yet detection-safe tokens.

\subsection{Detector-Aware White-Box Robustness}
\noindent\textbf{Experiment Setting.}
Beyond input-agnostic post-processing, we evaluate fake-to-real evasion using PGD and FGSM in token-embedding space. Both attacks use \(\epsilon=0.01\), and we report average F1 over all 29 held-out entries for CLIP, UNITE, and \methodname.

\begin{table}[H]
\centering
\small
\renewcommand{\arraystretch}{0.9}
\begin{tabular}{lccc}
\toprule
Attack & CLIP & UNITE & \methodname \\
\midrule
PGD & 0.249 & 0.182 & \textbf{0.501} \\
FGSM & 0.495 & 0.213 & \textbf{0.583} \\
\bottomrule
\end{tabular}
\caption{F1 under white-box feature-space attacks with \(\epsilon=0.01\).}
\label{tab:whitebox_attacks}

\end{table}

\noindent\textbf{Experiment Result and Analysis.}
As shown in Table~\ref{tab:whitebox_attacks}, \methodname\ substantially outperforms both baselines under PGD and FGSM. Its PGD advantage is 25.2 points over CLIP and 31.9 points over UNITE, indicating that robustness extends beyond common visual corruptions to detector-aware gradient-based evasion.

\subsection{Codec-Normalized Evaluation}
\noindent\textbf{Experiment Setting.}
To control source-specific compression cues, we re-encode both real and generated videos with the same H.264 pipeline at CRF18, CRF23, and CRF28. We evaluate five representative OOD entries and report AUC; all other preprocessing and the trained detector remain fixed.

\begin{table}[H]
\centering
\scriptsize
\setlength{\tabcolsep}{2.6pt}
\renewcommand{\arraystretch}{0.9}
\begin{adjustbox}{max width=\columnwidth}
\begin{tabular}{lcccccc}
\toprule
CRF & MuSeV & OpenSora1.2 & EasyAnimate & CogVideoX & Wan2.6 & Avg. \\
\midrule
18 & 0.660 & 0.952 & 0.964 & 0.971 & 0.959 & \textbf{0.901} \\
23 & 0.638 & 0.950 & 0.948 & 0.957 & 0.963 & \textbf{0.891} \\
28 & 0.616 & 0.944 & 0.928 & 0.935 & 0.962 & \textbf{0.877} \\
Range (pp) & 4.40 & 0.78 & 3.57 & 3.55 & 0.35 & \textbf{2.53} \\
\bottomrule
\end{tabular}
\end{adjustbox}
\caption{Codec-normalized AUC under matched H.264 re-encoding.}
\label{tab:codec_normalized}

\end{table}

\noindent\textbf{Experiment Result and Analysis.}
Table~\ref{tab:codec_normalized} reports average AUCs of \(0.901\), \(0.891\), and \(0.877\) from CRF18 to CRF28, with a mean cross-CRF range of only 2.53 points. The stable results across matched re-encoding settings suggest that \methodname\ does not primarily rely on source-level codec artifacts.

\subsection{Selected-Token Intervention}
\noindent\textbf{Experiment Setting.}
To characterize the selected tokens without imposing an adversarial direction, we replace them with the video's average token embedding and measure generator-ID accuracy and real/fake AUC before and after intervention.

\begin{table}[H]
\centering
\small
\renewcommand{\arraystretch}{0.9}
\begin{tabular}{lcc}
\toprule
Representation & \makecell{Generator-ID\\Acc} & \makecell{Real/Fake\\AUC} \\
\midrule
Original & 93.75 & 99.94 \\
Selected tokens replaced & 82.25 & 99.50 \\
\bottomrule
\end{tabular}
\caption{Neutral intervention on tokens selected by \methodname.}
\label{tab:token_intervention}

\end{table}

\noindent\textbf{Experiment Result and Analysis.}
Replacing selected tokens reduces generator-ID accuracy by 11.50 points but changes real/fake AUC by only 0.44 points. These tokens are therefore enriched in generator-specific information while remaining largely non-critical to detection ranking, directly supporting the generator-sensitive yet detection-safe selection criterion.

\subsection{Interpretation and Scope}
\methodname\ is a training framework rather than a backbone-specific detector: it combines generator-versus-detection token selection with generator-ID-oriented embedding perturbation. The KL divergence in the perturbation update is computed over the generator-classifier output of the jointly perturbed representation; gradients with respect to each selected token then define token-specific directions while retaining self-attention interactions. The selected tokens are model-dependent forensic cues and need not correspond to one human-interpretable artifact type. The complete baseline results in the main paper show that every compared detector has a larger ID--OOD F1 gap than \methodname's 6.97 points. Remaining limitations include scalability to longer videos and broader real-world domains, which motivate future evaluation beyond the current benchmark.

\section{Complete Training Procedure}
\label{app:training_algorithm}

The complete alternating optimization procedure used to train \methodname\ is summarized in Algorithm~\ref{alg:forge_training}.

\begin{algorithm}[H]
\scriptsize
\caption{Adversarial Training Procedure of \methodname}
\label{alg:forge_training}
\begin{algorithmic}[1]
    \State \textbf{Input:} Training set $D_{\mathrm{train}}$, backbone encoder $\mathcal{E}(\cdot)$, scoring network $\mathcal{S}_{\psi}(\cdot)$, real/fake head $C_{\mathrm{det}}(\cdot)$, generator-ID head $C_{\mathrm{gen}}(\cdot)$
    \State \textbf{Initialize:} epoch $\leftarrow 0$, global step $t \leftarrow 0$

    \While{$\mathrm{epoch} < \mathrm{epoch}_{\max}$}
        \For{each batch $\{(V_i, y_i, g_i)\}_{i=1}^{N}$ in $D_{\mathrm{train}}$}

            \State Update balancing coefficient $\alpha_t \leftarrow \min\!\left(\alpha_{\max}, \frac{t}{T_{\mathrm{ramp}}}\alpha_{\max}\right)$

            \State \textbf{// Step 1: update attacker while freezing detector}
            \For{each sample $(V_i, y_i, g_i)$ in the batch}
                \State Obtain clean token representations $\mathbf{H}_i=\mathcal{E}(V_i)$
                \State Compute generator relevance score $s_i^{\mathrm{gen}}$
                \State Compute detection relevance score $s_i^{\mathrm{det}}$
                \State Compute refined perturbation priority $s_i^{\star}$
                \State Predict token priority $\hat{s}_i=\mathcal{S}_{\psi}(\mathbf{H}_i)$
            \EndFor
            \State Compute attacker loss $\mathcal{L}_{\mathrm{att}}$
            \State Update $\mathcal{S}_{\psi}$ using $\mathcal{L}_{\mathrm{att}}$ while keeping $\mathcal{E}$, $C_{\mathrm{det}}$, and $C_{\mathrm{gen}}$ frozen

            \State \textbf{// Step 2: construct perturbed representations}
            \State Initialize the perturbed batch set $\tilde{D} \leftarrow \emptyset$
            \For{each sample $(V_i, y_i, g_i)$ in the batch}
                \State Obtain clean token representations $\mathbf{H}_i=\mathcal{E}(V_i)$
                \State Predict token priority $\hat{s}_i=\mathcal{S}_{\psi}(\mathbf{H}_i)$
                \State Select the top-$m$ important token set $\mathcal{I}_i$
                \State Apply gradient-guided embedding perturbation and obtain $\tilde{\mathbf H}_i$
                \State Update $\tilde{D} \leftarrow \tilde{D} \cup \{(\mathbf{H}_i,\tilde{\mathbf H}_i,y_i,g_i)\}$
            \EndFor

            \State \textbf{// Step 3: update detector while freezing attacker}
            \State Compute generator-ID loss $\mathcal{L}_{\mathrm{ID}}$ on clean representations
            \State Compute real/fake detection loss $\mathcal{L}_{\mathrm{RFDL}}$ on clean and perturbed representations
            \State Compute forensic consistency loss $\mathcal{L}_{\mathrm{FCL}}$
            \State Compute detector loss $\mathcal{L}_{\mathrm{total}}$
            \State Update $C_{\mathrm{gen}}$ using $\mathcal{L}_{\mathrm{ID}}$
            \State Update $\mathcal{E}$ and $C_{\mathrm{det}}$ using $\mathcal{L}_{\mathrm{total}}$ while keeping $\mathcal{S}_{\psi}$ frozen

            \State $t \leftarrow t + 1$
        \EndFor
        \State $\mathrm{epoch} \leftarrow \mathrm{epoch} + 1$
    \EndWhile

    \State \textbf{Output:} Trained scorer $\mathcal{S}_{\psi}$ and detector $\mathcal{D}$, including $\mathcal{E}$, $C_{\mathrm{det}}$, and $C_{\mathrm{gen}}$
\end{algorithmic}
\end{algorithm}

\end{document}